\PassOptionsToPackage{table}{xcolor}
\documentclass[10pt]{article}

\usepackage[margin=1in]{geometry}
\usepackage{times}
\usepackage{graphicx}
\usepackage[hyphens]{url}
\usepackage{booktabs}
\usepackage{xcolor}
\usepackage{subcaption}
\usepackage{multirow}
\usepackage{colortbl}
\usepackage{caption}
\usepackage{amsmath}
\usepackage{amssymb}
\usepackage{algorithm}
\usepackage{algorithmic}
\usepackage{float}
\usepackage{makecell}
\usepackage{amsthm}
\usepackage[numbers]{natbib}

\usepackage{hyperref}
\hypersetup{
    colorlinks=true,
    citecolor=blue,
    linkcolor=blue,
    urlcolor=blue
}

\definecolor{oursrow}{RGB}{220,239,239}
\definecolor{sandwichrow}{RGB}{224,241,240}

\newcommand{\sandwichquant}{\textsc{SandwichQuant}}
\newcommand{\sqpost}{\textsc{SandwichQuant-Post}}
\newcommand{\sqpre}{\textsc{SandwichQuant-Pre}}

\newcommand{\safeincludegraphics}[2][]{%
  \IfFileExists{#2}{\includegraphics[#1]{#2}}{%
    \fbox{\parbox[c][1.15in][c]{0.94\linewidth}{\centering
    Figure placeholder\\\texttt{\detokenize{#2}}}}}}

\begin{document}
\title{SandwichQuant: Which Parameters Matter Before and After Quantization?}

\author{
Peng Xia \qquad Junbiao Pang\thanks{Corresponding author.}\\
School of Information Science and Technology, Beijing University of Technology\\
Beijing, China\\
\texttt{\{xiapeng@emails.bjut.edu.cn, junbiao\_pang@bjut.edu.cn\}}
}

\maketitle
\begin{abstract}
Quantization correction methods usually optimize weights, quantization parameters, or reconstruction objectives, while the underlying parameter subspaces responsible for effective correction remain unclear. In this work, we study quantization correction from a parameter subspace perspective and reveal that correction capability is highly non-uniform across parameter groups. By decomposing trainable parameters into backbone weights, normalization-affine parameters, and quantization parameters, we show that the low-dimensional normalization-affine subspace provides a highly efficient correction direction under matched budgets. Based on this finding, we propose SandwichQuant, a two-stage normalization-affine correction framework that performs adaptation before and after quantization. The pre-stage improves quantization robustness, while the post-stage compensates residual errors after the quantized graph is fixed. Extensive experiments on vision models and large language models demonstrate consistent improvements under various low-bit quantization settings, validating the effectiveness of subspace-aligned correction.

\end{abstract}

\section{Introduction}

Large language models (LLMs) have achieved remarkable performance across
language understanding, generation, and reasoning, but their growing parameter
counts make inference expensive in both memory and computation. Quantization
alleviates this cost by mapping floating-point weights and activations to
low-precision representations. Quantization-aware training (QAT) adapts a model
under simulated low-precision arithmetic and often provides strong accuracy,
but its training cost is increasingly prohibitive at LLM scale
~\cite{jacob2018quantization,esser2020lsq,bhalgat2020lsqplus}. Post-training
quantization (PTQ) is therefore the dominant deployment paradigm: it calibrates
a pretrained model using limited data, without full retraining
~\cite{adaround,brecq,qdrop,gptq,awq,spinquant}. Nevertheless, accuracy still
degrades sharply at extreme precision, especially when weights, activations,
and KV caches are quantized jointly. Improving this regime is commonly treated
as a problem of finding better rounding decisions, quantization grids,
transformations, or weight-compensation rules.

We expose a complementary source of robustness that remains largely hidden by
this view. Once a quantizer has fixed its weights, rounding decisions, scales,
zero points, and rotations, updating only the affine parameters of
normalization layers can recover a large fraction of the remaining functional
error. The same affine coordinates can also be adapted before PTQ, changing the
response distribution presented to calibration. This raises a more basic
question than how to design another quantizer: \emph{which parameter subspaces
matter for correction, and should they act before PTQ, after PTQ, or on both
sides?}

To answer this question, we decompose the trainable state as
$\Theta=(W,\Phi,\Omega)$, where $W$ denotes backbone weights, $\Phi$
normalization-affine parameters, and $\Omega$ quantizer parameters. We compare
these subspaces under the same target quantized graph and matched optimization
budgets, and study them through response analysis, controlled interventions,
and task-aware error geometry. The evidence reveals
that correction leverage is highly non-uniform: although $\Phi$ is orders of
magnitude smaller than $W$, each affine parameter broadcasts a channel-wise
transformation across all tokens or spatial locations and can therefore
redirect the response trajectory of many downstream layers. This leverage is
powerful but not unlimited. Affine adaptation can repair structured response
mismatch and error propagation, but it cannot recreate information destroyed
by severe clipping or rounding. Thus, functional recovery depends not simply
on parameter count or feature reconstruction, but on whether the available
subspace is aligned with the recoverable component of quantization error.

This finding motivates \sandwichquant. For a base backend $\mathcal B$, we
first learn $\Phi_{\mathrm{pre}}$, transfer only that affine state to the
original dense checkpoint, and rerun $\mathrm{PTQ}_{\mathcal B}$ from scratch.
We then freeze the completed low-bit graph and learn an independent
$\Phi_{\mathrm{post}}$. Thus
$\Phi_{\mathrm{pre}}\!\rightarrow\!\mathrm{PTQ}_{\mathcal B}
\!\rightarrow\!\Phi_{\mathrm{post}}$
separates distribution preconditioning from residual response correction.
Both stages modify parameters already present in the network, adding neither an
auxiliary inference branch nor a new deployment module. One-sided \sqpre{} and
\sqpost{} variants are retained as controlled components, rather than being
reported as the complete method. Existing normalization tuning methods~\cite{normtweaking} demonstrate that affine parameters can improve quantized representations. However, they mainly treat normalization adaptation as a practical correction heuristic. In contrast, we study normalization affine parameters as one candidate subspace among the entire parameter space, establish matched-budget evidence against equally-sized weight subspaces, and analyze why this subspace has high correction leverage.

Our main contributions are:
\begin{itemize}
    \item We formulate quantization correction over the coordinate groups
    $W$, $\Phi$, and $\Omega$, and show that, under the studied quantization settings, a substantial portion of recoverable quantization error can be addressed within the low-dimensional normalization-affine subspace.

    \item We explain this concentration through response propagation,
    same-budget subspace controls, functional interventions, and task-aware
    error geometry. The resulting picture separates
    correctable response mismatch from irreversible low-bit information loss.

    \item We introduce \sandwichquant{}, a two-sided affine framework that combines quantization preconditioning with frozen-graph post-correction. Across three LLM families, multiple PTQ backends, weight-only W3A16, and joint W2A4KV4, the complete pipeline consistently lowers perplexity and improves mean zero-shot accuracy without introducing an additional inference-time module or operator.
\end{itemize}

\section{Related Work}

\paragraph{Quantization correction and distribution transformation.}
Classical PTQ improves a fixed model through equalization, rounding, or local
reconstruction~\cite{nagel2019dfq,adaround,brecq,qdrop,gptq}. More recent work
models how local errors propagate across blocks~\cite{qep,qronos}, incorporates
Hessian, Fisher, or perturbation sensitivity~\cite{lsvit,aphqvit,fimaq}, or
introduces auxiliary and low-rank correction paths
~\cite{qwt,aiqvit,qsca,svdquant,firstordererror}. A complementary family
reshapes quantization difficulty through channel scaling, activation-aware
weight protection, rotations, or invertible transforms
~\cite{smoothquant,awq,spinquant,quarot,ostquant,dartquant,flatquant}. GPTAQ and
ResComp are particularly relevant because they compensate errors across
sequentially quantized layers~\cite{gptaq,rescomp}. These methods ask how to
improve a quantizer, transform, or reconstruction objective; we instead ask
which existing parameter subspaces provide high-leverage directions before
and after PTQ, and compose the two roles in \sandwichquant.

\paragraph{Normalization-based adaptation.}
Norm Tweaking updates normalization parameters to realign quantized and
floating-point LLM activations~\cite{normtweaking}, while PTQ4VM uses global
affine calibration to correct accumulated distortion associated with Batch
Normalization~\cite{ptq4vm}. More broadly, normalization has been connected to
smoother optimization and reduced sharpness~\cite{bnoptimization,normsharpness};
training only normalization variables can retain substantial expressive power
or support test-time adaptation~\cite{trainbnonly,tent}. These results establish
the practical value of normalization-related variables. We instead compare
$\Phi$ with equally sized weight-coordinate subspaces under shared calibration
and optimization budgets, and analyze where affine leverage concentrates and
whether feature matching is necessary for functional recovery.

\paragraph{QAT, flatness, and low-dimensional subspaces.}
QAT adapts a model directly under fake quantization
~\cite{esser2020lsq,bhalgat2020lsqplus}; oscillation-aware and ultra-low-bit
variants further stabilize aggressive quantization
~\cite{qat_oscillation,reasonqat,arbitraryprecision}. Checkpoint trajectories
and spectral conditioning also affect PTQ robustness
~\cite{trainingdynamics,s2d}. This connects to sharp minima and
perturbation-robust optimization~\cite{keskar,dinh,sam,asam,relativeflatness,swa};
notably, normalization-restricted SAM retains much of the benefit of
full-parameter perturbations~\cite{samnorm}. Intrinsic-dimension studies and
parameter-efficient adaptation likewise show that useful updates may occupy
small subspaces~\cite{intrinsicdimension,intrinsicfinetuning,tinysubspace,
adapter,bitfit,lora}. Our setting differs because quantization noise is
structured and anisotropic: we measure subspace leverage under the actual
rounding, clipping, and activation-quantization graph rather than generic
isotropic parameter perturbations.

\section{Method}
\label{sec:method}

\subsection{Parameter-Subspace Formulation}
\label{sec:overview}

\begin{figure*}[t]
\centering

\begin{subfigure}[t]{0.485\textwidth}
    \centering
    \safeincludegraphics[width=\linewidth]{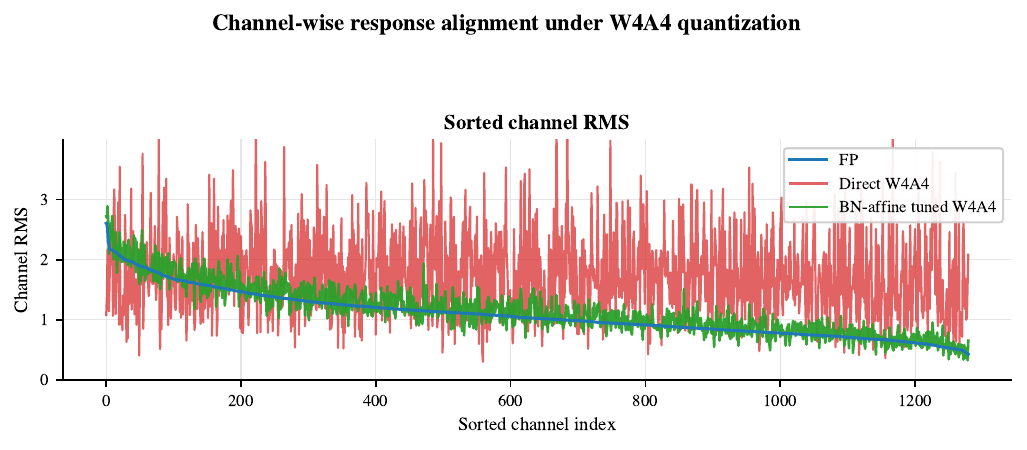}
    \caption{
    \textbf{Response control under W4A4.}
    BN-affine adaptation substantially reduces channel-wise RMS mismatch.
    }
    \label{fig:feature_alignment}
\end{subfigure}
\hfill
\begin{subfigure}[t]{0.485\textwidth}
    \centering
    \safeincludegraphics[width=\linewidth]{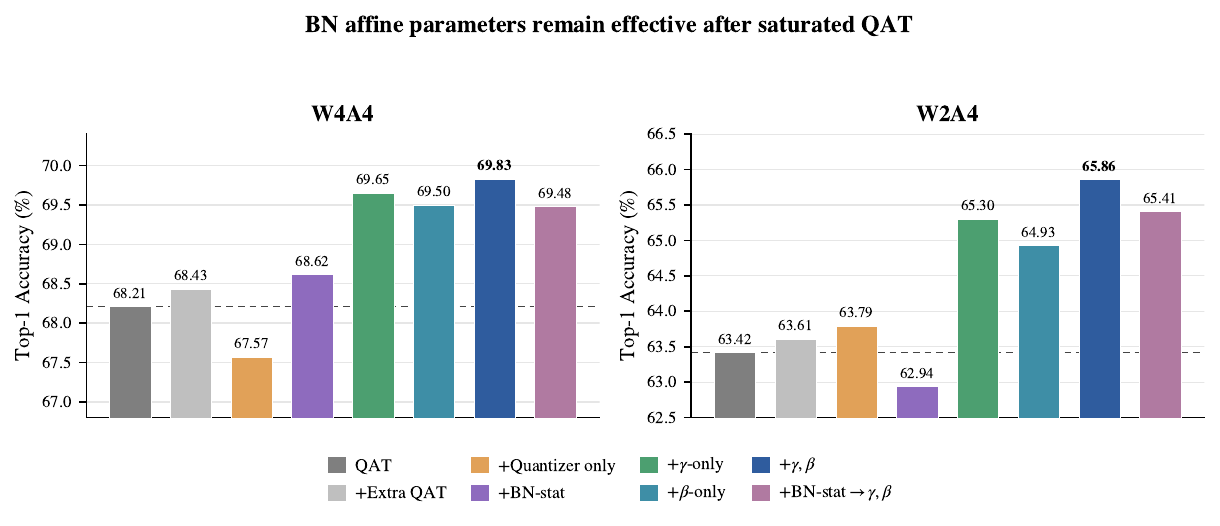}
    \caption{
    \textbf{Residual directions after QAT.}
    Affine-only refinement remains useful from the same LSQ+ checkpoint.
    }
    \label{fig:qat_diagnostic}
\end{subfigure}

\caption{
\textbf{A low-dimensional affine subspace provides high-leverage correction
directions before and after full-space optimization.}
The post-QAT result diagnoses useful residual directions under a fixed protocol;
it does not imply greater asymptotic capacity than full-space training.
}
\label{fig:affine_overview}
\end{figure*}

Figure~\ref{fig:affine_overview} establishes the cross-regime phenomenon that
motivates our method. The same structured affine coordinates remain useful both
before a PTQ reconstruction stage and after a converged QAT checkpoint. These
controls do not by themselves constitute \sandwichquant; they show that
$\mathcal S_{\Phi}$ contains useful directions on both sides of quantization.

Let $f(\boldsymbol{x};\mathbf{W},\boldsymbol{\Phi})$ denote a pretrained
network and let $\boldsymbol{\Omega}$ collect the parameters of the target
fake-quantization graph. We decompose the trainable state as
\begin{equation}
\Theta=(\mathbf{W},\boldsymbol{\Phi},\boldsymbol{\Omega}),
\end{equation}
where $\mathbf{W}$ contains convolutional and linear weights/biases,
$\boldsymbol{\Phi}$ contains normalization-affine parameters, and
$\boldsymbol{\Omega}$ contains quantizer variables. For BN~\cite{batchnorm},
$\boldsymbol{\Phi}$ contains channel-wise $(\boldsymbol{\gamma},
\boldsymbol{\beta})$; for RMSNorm~\cite{zhang2019rmsnorm}, only
$\boldsymbol{\gamma}$ is present.

We define the corresponding coordinate subspaces by restricting an update
$\Delta\Theta$ to one parameter group, e.g.,
\begin{equation}
\mathcal{S}_{\Phi}
=
\{\Delta\Theta:\Delta\mathbf{W}=0,\;\Delta\boldsymbol{\Omega}=0\},
\label{eq:phi_subspace}
\end{equation}
with analogous definitions for $\mathcal{S}_{W}$ and
$\mathcal{S}_{\Omega}$. The purpose of the controlled subspace study is to
compare the correction achieved by these structured update directions under the
same quantized initialization and optimization budget. Our affine stages
specialize to $\mathcal{S}_{\Phi}$ and optimize only existing response-control
coordinates. \sqpre{} acts before the base backend, whereas \sqpost{} keeps
the completed feature extractor and quantization grid fixed.

\subsection{Target-Aligned Affine Stages and SandwichQuant}
\label{sec:sandwichquant}

Let $\mathcal B_{\mathcal D}$ denote a complete PTQ backend calibrated on
$\mathcal D$, including rounding, clipping, rotations, reconstruction, and
compensation.  We write its state transition as
\begin{equation}
\mathcal B_{\mathcal D}(\mathbf W,\boldsymbol\Phi)
\mapsto
\mathcal G=(\widehat{\mathbf W},\boldsymbol\Phi,
\boldsymbol\Omega_{\mathcal B}),
\label{eq:backend_operator}
\end{equation}
and define $\mathcal A_T(\mathcal G;\boldsymbol\Phi_{\rm init})$ as $T$ steps
of Eq.~\eqref{eq:sandwich_objective} over $\boldsymbol\Phi$ only, starting at
$\boldsymbol\Phi_{\rm init}$ while every other tensor and the graph topology
remain fixed.

\paragraph{Complete construction.}
\sandwichquant{} applies this affine operator on both sides of the deployable
PTQ graph:
\begin{subequations}\label{eq:sandwich_pipeline}
\begin{align}
\mathcal G_1
&=\mathcal B_{\mathcal D}(\mathbf W_0,\boldsymbol\Phi_0),
&
\boldsymbol\Phi_{\rm pre}
&=\mathcal A_{T_{\rm pre}}(\mathcal G_1;\boldsymbol\Phi_0),
\label{eq:sandwich_pre}\\
\mathcal G_2
&=\mathcal B_{\mathcal D}(\mathbf W_0,\boldsymbol\Phi_{\rm pre}),
&
\boldsymbol\Phi_{\rm post}
&=\mathcal A_{T_{\rm post}}(\mathcal G_2;\boldsymbol\Phi_{\rm pre}),
\label{eq:sandwich_post}\\
\mathcal G_{\rm SQ}
&=\mathcal G_2[\boldsymbol\Phi\leftarrow
\boldsymbol\Phi_{\rm post}].
\label{eq:sandwich_deploy}
\end{align}
\end{subequations}
The first graph $\mathcal G_1$ is a \emph{disposable target-backend probe}.
After learning $\boldsymbol\Phi_{\rm pre}$, we reset the original dense
checkpoint and transfer only these affine tensors; quantized weights,
quantizer variables, and optimizer state from $\mathcal G_1$ are discarded.
The backend is then rerun from scratch on the same calibration tensor to form
$\mathcal G_2$.  The post stage starts from the affine state carried by
$\mathcal G_2$ with a fresh optimizer and corrects only its residual response.
Thus \sqpre{} changes the responses presented to the backend and can alter its
quantized solution, whereas \sqpost{} selects a better affine state on one fixed
solution.  Their composition is not equivalent to extending either one-sided
stage.

After calibration on $\mathcal{D}_{c}$, the quantization parameters are fixed
as $\boldsymbol{\Omega}_{0}$. The target fake-quantized student and the fixed
full-precision teacher are
\begin{align}
\boldsymbol{z}_{q}
&=
\widetilde{f}_{q}
\left(
\boldsymbol{x};
\mathbf{W}_{0},
\boldsymbol{\Phi},
\boldsymbol{\Omega}_{0},
\mathcal{G}_{\mathcal{B}}
\right),
\label{eq:student}\\
\boldsymbol{z}_{0}
&=
f
\left(
\boldsymbol{x};
\mathbf{W}_{0},
\boldsymbol{\Phi}_{0}
\right).
\label{eq:teacher}
\end{align}
The affine stage freezes convolutional and linear weights and updates only the
normalization-affine parameters:
\begin{equation}
\nabla_{\mathbf{W}}\mathcal{L}_{\mathrm{SQ}}=\mathbf{0},
\qquad
\nabla_{\boldsymbol{\Phi}}\mathcal{L}_{\mathrm{SQ}}\neq\mathbf{0}.
\label{eq:freeze_backbone}
\end{equation}

\begin{figure*}[t]
    \centering
    \includegraphics[
        width=0.9\textwidth,
        trim=0 0 0 0,
        clip
    ]{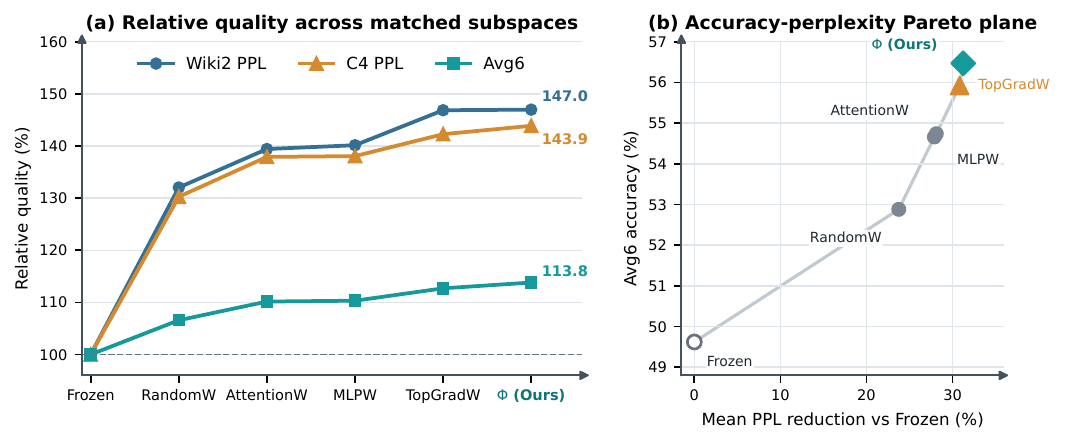}
    \caption{
        \textbf{Matched-budget parameter-subspace comparison on Qwen3-8B
        under W2A4KV4.}
        All trainable alternatives use exactly the same number of scalar
        parameters as the normalization-affine subspace $\Phi$ and share
        identical calibration data, optimization steps, and training
        objectives.
        \textbf{(a)} Relative WikiText-2 perplexity, C4 perplexity, and
        Avg6 accuracy, normalized to the frozen quantized baseline
        (higher is better).
        \textbf{(b)} Accuracy--perplexity trade-off, where the horizontal
        axis reports the mean relative reduction in WikiText-2 and C4
        perplexity against the frozen baseline.
        Despite its extremely low dimensionality, $\Phi$ achieves the
        strongest overall recovery and slightly outperforms the
        gradient-selected TopGradW subspace.
    }
    \label{fig:matched_subspace_qwen3}
\end{figure*}

\begin{figure*}[t]
    \centering

    \begin{subfigure}[t]{0.495\textwidth}
        \centering
        \includegraphics[width=\linewidth]
        {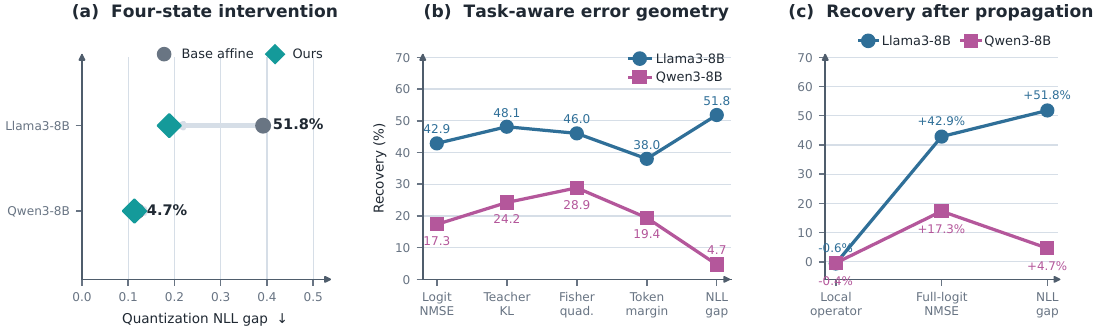}
        \caption{\textbf{Weight-only W3A16.}
        Mechanism analysis on Llama3-8B and Qwen3-8B under GPTQ.
        The one-sided \sqpost{} probe contracts the quantization-induced NLL
        gap and consistently improves task-aware error geometry.}
        \label{fig:mechanism_w3a16}
    \end{subfigure}
    \hfill
    \begin{subfigure}[t]{0.495\textwidth}
        \centering
        \includegraphics[width=\linewidth]
        {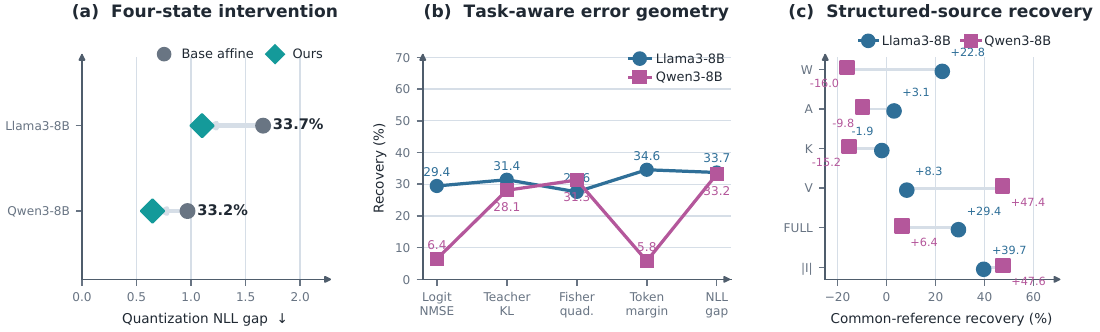}
        \caption{\textbf{Joint W2A4KV4.}
        Mechanism analysis under QuaRot+ResComp.
        Recovery remains substantial under simultaneous weight, activation,
        and KV-cache quantization, while the structured-source intervention
        reveals non-uniform and non-additive error correction.}
        \label{fig:mechanism_w2a4kv4}
    \end{subfigure}

    \caption{\textbf{The post-PTQ affine component reshapes the functional
    geometry of quantization error across two regimes.}
    Each subfigure reports the four-state intervention, task-aware recovery
    metrics, and structured-source analysis for Llama3-8B and Qwen3-8B.
    These one-sided probes isolate \sqpost{}, the residual-correction component
    of \sandwichquant. Across weight-only W3A16 and joint W2A4KV4, it reduces
    task-relevant degradation more consistently than isolated local operator
    error, indicating that its principal effect emerges through propagation and
    interaction rather than uniform local reconstruction.}
\label{fig:mechanism_joint}
\end{figure*}

Figure~\ref{fig:matched_subspace_qwen3} rules out the explanation that any
equal-size set of trainable scalars is sufficient: on Qwen3-8B W2A4KV4,
$\mathcal S_{\Phi}$ attains the best joint perplexity--accuracy trade-off and
slightly exceeds the gradient-selected weight control. The mechanism probes in
Figure~\ref{fig:mechanism_joint} then clarify what this subspace changes.
\sqpost{} reduces the quantization-induced NLL gap for both model families,
while task-aware recovery can be much larger than recovery of an isolated local
operator. The structured-source interventions are non-additive, supporting a
propagation-control interpretation rather than uniform layerwise reconstruction.

We use a task loss together with prediction-level knowledge
distillation~\cite{hinton2015distilling}. The task loss
is classification cross-entropy for image classification, pixel-wise
cross-entropy for semantic segmentation, and next-token negative
log-likelihood for language modeling. Let
$p_T(\boldsymbol{z})=\operatorname{softmax}(\boldsymbol{z}/T)$. The \sandwichquant{}
objective is
\begin{equation}
\mathcal{L}_{\mathrm{SQ}}
=
\mathcal{L}_{\mathrm{task}}(\boldsymbol{z}_{q},y)
+
\lambda_{\mathrm{KD}}T^2
\operatorname{KL}
\left(
p_T(\boldsymbol{z}_{0})
\Vert
p_T(\boldsymbol{z}_{q})
\right).
\label{eq:sandwich_objective}
\end{equation}
The task term keeps the fake-quantized model predictive, while the
distillation term preserves the decision structure of the fixed teacher.
Crucially, the loss is optimized on the same target graph used for evaluation
or downstream reconstruction. Changing that graph changes both the residual
and the directions through which $\boldsymbol\Phi$ can correct it; an affine
checkpoint is therefore not expected to transfer across mismatched backends.

\begin{algorithm}[t]
\caption{\sandwichquant{} with base PTQ backend $\mathcal B$}
\label{alg:sandwichquant}
\begin{algorithmic}[1]
\REQUIRE Dense student state $(\mathbf W_0,\boldsymbol\Phi_0)$; frozen dense
teacher $f_T$; PTQ backend $\mathcal B$; shared calibration tensor $\mathcal D$;
affine budgets $T_{\rm pre}$ and $T_{\rm post}$
\ENSURE Frozen deployable graph $\mathcal G_{\rm SQ}$
\STATE Save an immutable dense snapshot
$\mathcal S_0\leftarrow(\mathbf W_0,\boldsymbol\Phi_0)$
\STATE Build the disposable graph $\mathcal G_1$ using the first relation in
Eq.~\eqref{eq:sandwich_pre}
\STATE Freeze weights and quantizer variables in $\mathcal G_1$; enable
gradients only for $\boldsymbol\Phi$
\STATE Obtain $\boldsymbol\Phi_{\rm pre}$ from the second relation in
Eq.~\eqref{eq:sandwich_pre}, optimizing Eq.~\eqref{eq:sandwich_objective}
\STATE Discard $\mathcal G_1$ and its optimizer state; restore $\mathcal S_0$
\STATE Transfer only $\boldsymbol\Phi_{\rm pre}$ into the restored dense state
\STATE Rebuild PTQ from scratch to obtain $\mathcal G_2$ using the first
relation in Eq.~\eqref{eq:sandwich_post}
\STATE Freeze the completed graph $\mathcal G_2$ except for
$\boldsymbol\Phi$
\STATE Obtain $\boldsymbol\Phi_{\rm post}$ from the second relation in
Eq.~\eqref{eq:sandwich_post}, optimizing Eq.~\eqref{eq:sandwich_objective}
\STATE Install $\boldsymbol\Phi_{\rm post}$, freeze all parameters, and record
the backend, calibration, and artifact identifiers
\STATE \textbf{return} the deployable graph $\mathcal G_{\rm SQ}$ defined in
Eq.~\eqref{eq:sandwich_deploy}
\end{algorithmic}
\end{algorithm}

Algorithm~\ref{alg:sandwichquant} changes construction-time state but adds no inference
branch. BN affine factors can be folded into an adjacent convolution, and
RMSNorm scales are already native model parameters. The two affine stages and
the extra PTQ pass add offline work; optimizer state is maintained only for
$\boldsymbol\Phi$, although activation computation still contributes to
wall-clock time. Exact budgets and artifact-reset semantics are reported in
Appendix~\ref{app:llm_protocol}.

\long\def\deferredqatmethod{
\subsection{\sandwichquant{} after QAT and Alternating SandwichQuant-QAT}
\label{sec:sq_qat}

The same affine operator can also be applied to a saturated QAT checkpoint
$(\mathbf{W}_{\mathrm{qat}},\boldsymbol{\Phi}_{\mathrm{qat}},
\boldsymbol{\Omega}_{\mathrm{qat}})$. In this setting, the backbone and
quantization parameters are fixed and only the normalization affine
parameters are adapted:
\begin{equation}
\boldsymbol{\Phi}^{*}
=
\arg\min_{\boldsymbol{\Phi}}
\mathcal{L}_{\mathrm{SQ}}
\left(
\mathbf{W}_{\mathrm{qat}},
\boldsymbol{\Phi},
\boldsymbol{\Omega}_{\mathrm{qat}}
\right).
\label{eq:post_qat_sq}
\end{equation}
This post-QAT form tests whether a saturated jointly trained model still
contains an under-exploited affine adaptation direction.

For large language models, we additionally use an alternating variant that
decouples quantization-grid adaptation from response adaptation. Backbone
weights remain frozen. We denote by $\mathcal{L}_{q}$ the same
target-graph task-and-distillation objective when it is optimized with
respect to the quantization parameters. At alternating round $r$, we first
update group-wise quantization scales while fixing normalization parameters,
\begin{equation}
\boldsymbol{\Omega}^{r+1}
\leftarrow
\arg\min_{\boldsymbol{\Omega}}
\mathcal{L}_{q}
\left(
\mathbf{W}_{0},
\boldsymbol{\Phi}^{r},
\boldsymbol{\Omega}
\right),
\label{eq:qscale_step}
\end{equation}
and then update RMSNorm scales while fixing the quantization grid,
\begin{equation}
\boldsymbol{\Phi}^{r+1}
\leftarrow
\arg\min_{\boldsymbol{\Phi}}
\mathcal{L}_{\mathrm{SQ}}
\left(
\mathbf{W}_{0},
\boldsymbol{\Phi},
\boldsymbol{\Omega}^{r+1}
\right).
\label{eq:sq_step}
\end{equation}
We refer to this diagnostic block-coordinate extension as SandwichQuant-QAT. QScale-QAT adapts only
$\boldsymbol{\Omega}$, RMSNorm affine adaptation adapts only $\boldsymbol{\Phi}$, and
SandwichQuant-QAT alternates between the two complementary subspaces. This formulation
also clarifies that the large-model experiments do not perform
full-parameter QAT.

}

\paragraph{Extensions.}
The affine objective can also diagnose residual directions after QAT or be
alternated with quantizer-grid updates. These extensions are not part of the
headline \sandwichquant{} pipeline and are detailed in
Appendix~\ref{app:vision_qat}.

\long\def\deferredmechanism{
\subsection{Exact BN Scale--Shift Cancellation}
Assume fixed BN statistics $(\mu_c,\sigma_c)$ and a local pre-BN perturbation
\begin{equation}
\widehat a_c=\alpha_c a_c+\delta_c+e_c,
\qquad \alpha_c>0,
\label{eq:error_decomposition}
\end{equation}
where $e_c$ contains sample-dependent rounding, clipping, and saturation
residuals.  Let $\rho_c=\sqrt{\sigma_c^2+\epsilon}$ and let
$(\gamma_c^0,\beta_c^0)$ be the original BN affine parameters.  Choosing
\begin{align}
\gamma_c^*&=\gamma_c^0/\alpha_c,\\
\beta_c^*&=\beta_c^0-
\frac{\gamma_c^0}{\alpha_c\rho_c}
\big[(\alpha_c-1)\mu_c+\delta_c\big]
\label{eq:compensation}
\end{align}
gives
\begin{equation}
\widehat y_c^*=y_c^0+
\frac{\gamma_c^0}{\alpha_c\rho_c}e_c,
\qquad
\|\widehat y_c^*-y_c^0\|
\leq
\frac{|\gamma_c^0|}{\alpha_c\rho_c}\|e_c\|.
\label{eq:compensation_residual}
\end{equation}
Hence the structured scale--shift component is exactly cancellable under these
assumptions, but the unstructured residual remains.  This identity is specific
to BN with fixed statistics. RMSNorm lacks a bias term and therefore supplies
multiplicative response control, not exact cancellation of a general additive
offset.

\subsection{Weighted Projection and Its Limits}
For the local problem in Eq.~\eqref{eq:projection}, a minimum-norm solution is
\begin{equation}
\Delta\boldsymbol\Phi^*=
(\mathbf J_{\Phi}^{\top}\mathbf H_z\mathbf J_{\Phi})^{\dagger}
\mathbf J_{\Phi}^{\top}\mathbf H_z\boldsymbol e_z.
\label{eq:projection_solution}
\end{equation}
Writing $\boldsymbol e_{\parallel}=\mathbf J_{\Phi}
\Delta\boldsymbol\Phi^*$ and
$\boldsymbol e_{\perp}=\boldsymbol e_z-\boldsymbol e_{\parallel}$ separates
the locally expressible component from the residual orthogonal in the
$\mathbf H_z$ geometry. This is a first-order characterization: finite-step
optimization of a nonlinear network need not compute the exact projector, and
neither component implies global flatness. Information removed by clipping or
rounding may lie outside the reachable response subspace and cannot in general
be reconstructed.
}

\subsection{Why Is the Affine Subspace High-Leverage?}
\label{sec:mechanism}
\paragraph{Broadcast response control.}
For normalized channel $c$ at layer $l$,
\begin{equation}
\boldsymbol a_{l,c}=\gamma_{l,c}\boldsymbol h_{l,c}+\beta_{l,c},
\qquad
\Delta\boldsymbol a_{l,c}=
\boldsymbol h_{l,c}\Delta\gamma_{l,c}+
\mathbf 1\Delta\beta_{l,c},
\label{eq:broadcast_update}
\end{equation}
with $\beta_{l,c}=0$ for RMSNorm. One or two scalars therefore alter an entire
channel over all tokens or spatial locations and every downstream consumer.
This broadcast structure explains how a very small coordinate set can have a
large functional footprint. In a local BN model it can cancel channel-wise
scale and shift exactly, while RMSNorm provides multiplicative gain control;
the assumptions and residual bound are given in
Appendix~\ref{app:mechanism_derivation}.

\paragraph{Correctable residual geometry.}
Let $\boldsymbol e_z=\boldsymbol z_0-
\boldsymbol z_q(\boldsymbol\Phi_0)$ and let $\mathbf J_{\Phi}$ be the logit
Jacobian of the frozen target graph. For a small affine update,
\begin{equation}
\boldsymbol z_q(\boldsymbol\Phi_0+\Delta\boldsymbol\Phi)
\approx \boldsymbol z_q(\boldsymbol\Phi_0)+
\mathbf J_{\Phi}\Delta\boldsymbol\Phi,
\end{equation}
so the local task-and-distillation objective induces
\begin{equation}
\min_{\Delta\boldsymbol\Phi}
\|\boldsymbol e_z-\mathbf J_{\Phi}\Delta\boldsymbol\Phi\|_{\mathbf H_z}^2,
\label{eq:projection}
\end{equation}
where $\mathbf H_z\succeq0$ is the local output-space curvature.  Thus
$\boldsymbol e_z=\boldsymbol e_{\parallel}+\boldsymbol e_{\perp}$ separates
an affine-expressible component from a locally irrecoverable component.  This
does not claim that finite-step nonlinear training is an exact orthogonal
projection; it predicts only that task-relevant residuals aligned with
$\operatorname{col}(\mathbf J_{\Phi})$ can be corrected efficiently. The
matched-budget and structured-source evidence in
Figures~\ref{fig:matched_subspace_qwen3}--\ref{fig:mechanism_joint} tests this
prediction directly.

\section{Experiments}
\label{sec:experiments}

\paragraph{Models, quantization, and calibration.}
We evaluate Llama2-7B~\cite{llama2}, Llama3-8B~\cite{llama3}, and
Qwen3-8B~\cite{qwen3}. Weight-only W3A16 uses symmetric group-wise
quantization with group size 128 and 128 C4 training sequences of length 2,048
(seed 0). Joint W2A4KV4 uses symmetric clipped W2 weights with group size 128,
dynamic per-token asymmetric A4/K4/V4 quantization, and 128 WikiText-2 training
sequences of length 2,048 (seed 0). Its QuaRot backends use Hadamard rotations;
GPTQ-family updates use damping 0.01 and activation ordering. Within each
model--regime pair, all methods share the dense checkpoint, tokenizer,
calibration indices, quantization configuration, and evaluation code.

\paragraph{Affine stages and evaluation.}
Each pre- or post-stage updates only normalization-affine tensors for 200 AdamW
steps (batch size 1, sequence length 256, learning rate $5\times10^{-4}$, zero
weight decay, gradient clipping 1.0) with equal-weight cross-entropy and
full-vocabulary teacher distillation at temperature 1. Between stages, PTQ is
rebuilt from the original dense checkpoint on the same calibration tensor; no
quantized weight, rounding decision, or quantizer parameter is transferred.
We report WikiText-2/C4 perplexity and the unweighted mean zero-shot accuracy
over PIQA, ARC-Easy, ARC-Challenge, HellaSwag, WinoGrande, and BoolQ. Vision
and QAT experiments serve as cross-domain component controls. Complete LLM
and vision protocols are in Appendices~\ref{app:llm_protocol}
and~\ref{app:vision_qat}, respectively.

\long\def\deferredvisionexperiments{
\subsection{Experimental Protocol}
\label{sec:setup}

\paragraph{Tasks and models.}
We evaluate \sandwichquant{} in four settings. We use
MobileNetV2~\cite{sandler2018mobilenetv2} on
ImageNet-1K~\cite{deng2009imagenet} for extreme weight--activation PTQ,
MobileNetV2 QAT checkpoints on CIFAR-100~\cite{krizhevsky2009cifar} for
post-QAT adaptation, and an ImageNet-pretrained ResNet34-based
U-Net~\cite{he2016resnet,ronneberger2015unet} on
Cityscapes~\cite{cordts2016cityscapes} for semantic segmentation. We further
evaluate Qwen2.5-3B-Instruct~\cite{qwen2024qwen25} on held-out
OSCAR~\cite{ortizsuarez2020oscar}, C4~\cite{raffel2020t5}, and
WikiText-2~\cite{merity2016wikitext}. We report ImageNet and CIFAR-100
top-1 accuracy, Cityscapes mIoU, and language-model perplexity.

\paragraph{Quantization settings.}
The CNN experiments are implemented with MQBench
~\cite{li2021mqbench}. Weights use symmetric per-output-channel
quantization. The MQBench Academic backend uses MSE observers and
\texttt{FixedFakeQuantize}; activations are quantized asymmetrically and
per-tensor, while the lightweight standalone graph uses unsigned activation
quantization after ReLU/ReLU6. The reported ImageNet results quantize the full
network, whereas the CIFAR-100 QAT diagnostic keeps the first and last layers
at 8 bits. For Qwen2.5-3B-Instruct, weights use symmetric group-wise linear
quantization with group size 128, activations use symmetric per-token
quantization, and the language-model head is excluded. Because standard AWQ
is weight-only~\cite{awq}, we use \emph{AWQ-style} for experiments
that combine activation-aware channel scaling with additional activation
quantization.

\paragraph{Strict native-subspace protocol.}
For the controlled W4A4 MobileNetV2/CIFAR-100 study, the first and last
quantized layers use 8 bits. A frozen branch calibrates activation qparams once
from 1,024 samples and saves them; all $W$, $\Phi$, $\Omega$,
$\Phi+\Omega$, and Full branches reload exactly the same calibration state.
The optimization set contains 40K samples, a fixed 4K split is used for model
selection, and the complete 10K test set is used only for the reported test
metrics. Every trainable branch uses the same CE+KD objective, AdamW with
learning rate $10^{-4}$ and zero weight decay, cosine annealing for 20 epochs,
$T=4$, $\lambda_{\mathrm{KD}}=1$, and gradient-norm clipping at 1.0.
$\Omega$ contains learnable per-output-channel weight scales and per-module
activation scales; activation zero-points remain fixed. This protocol is
intended to measure \emph{matched-budget correction leverage}, not separately
tuned asymptotic optima for each subspace.

\paragraph{Large-model optimization.}
For the Qwen2.5-3B-Instruct W4A4 experiment, all adaptation methods use 128
OSCAR documents for calibration and 1,920 disjoint documents for tuning.
Training sequences have length 256 and batch size 1. AWQ-style equalization, inspired by
AWQ~\cite{awq}, uses $\alpha=0.5$ and 16 calibration batches.
RMSNorm affine adaptation, QScale-QAT, and SandwichQuant-QAT use AdamW for at most 200 steps with
learning rate $5\times10^{-4}$, zero weight decay, $T=1$,
$\lambda_{\mathrm{CE}}=1$, $\lambda_{\mathrm{KD}}=1$, maximum gradient norm
1.0, and automatic mixed precision. The channel-scale coefficient is
$10^{-4}$ for RMSNorm affine adaptation and SandwichQuant-QAT and zero for QScale-QAT. SandwichQuant-QAT
alternates 10 QScale steps and 10 RMSNorm steps while keeping all backbone
weights frozen. Perplexity is computed on 128 held-out blocks of length 2,048
for each corpus.

For the weight-only extension, we evaluate Llama-2-7B, Llama-3-8B, and
Qwen3-8B under symmetric group-wise W3A16 quantization with group size 128.
Every method receives exactly the same 128 C4 calibration sequences of length
2,048, sampled with seed 0. \sandwichquant{} is applied only after the baseline quantizer is
frozen and updates the existing RMSNorm affine scales for at most 200 steps;
the quantized weights, quantizer parameters, and calibration tokens remain
unchanged. We report WikiText-2 and C4 perplexity, together with zero-shot
accuracy on PIQA, ARC-Easy, ARC-Challenge, HellaSwag, WinoGrande, and BoolQ.
For the first four tasks we use normalized accuracy; for WinoGrande and BoolQ
we use raw accuracy. The six-task average is the unweighted arithmetic mean.

\paragraph{Hardware, software, and runs.}
Experiments are implemented in PyTorch/MQBench or the corresponding standalone
fake-quantization graph. The strict shared-qparam subspace study was executed on
a single MetaX C500 GPU, while the local/block/feature-swap mechanism suite was
executed on MetaX C600-A hardware. The large-model experiments use MetaX C600-A
GPUs with the MACA PyTorch stack, Transformers 4.51.3, and eager attention for
perplexity evaluation. Unless otherwise stated, the current tables report the
seed-42 run.

\paragraph{Compared methods.}
RTN directly applies calibrated rounding without reconstruction.
QDrop~\cite{qdrop} is a reconstruction-based PTQ backend. \sandwichquant{} is
evaluated both as a standalone target-graph adaptation and as a
pre-conditioning stage before QDrop. For QAT checkpoints, we compare
PACT~\cite{choi2018pact}, DSQ~\cite{gong2019dsq},
LSQ~\cite{esser2020lsq}, LSQ+~\cite{bhalgat2020lsqplus},
oscillation dampening and iterative freezing
~\cite{qat_oscillation}, and StableQAT~\cite{chen2026stableqat}.
For language models, we compare RTN, AWQ-style scaling, RMSNorm affine adaptation,
QScale-QAT, and their alternating combination in the weight--activation study.
The weight-only study additionally compares GPTQ, GPTAQ, compensation-aware
GPTAQ (ResComp), and AWQ, each with and without the same post-quantization \sandwichquant{}
stage. QEP is retained as a diagnostic baseline in the appendix because its
gain is not consistent under our controlled implementation.

\subsection{Which Parameter Subspaces Matter for Quantization?}
\label{sec:native_subspaces}

We first test the central hypothesis directly, before evaluating \sandwichquant{} as a
method. Table~\ref{tab:native_subspaces} compares the native parameter groups
under the strict shared-qparam protocol described above. All trainable branches
start from the same calibrated W4A4 model (7.28\% top-1, up to a 0.01-point
numerical evaluation variation) and receive the same 20-epoch optimization
budget.

The result is strongly non-uniform. Although $\Omega$ has the highest raw gain
per trainable parameter, its absolute recovery saturates at 40.93\%. The
normalization-affine subspace reaches 60.19\% with only 34.2K parameters,
retaining 88.9\% of the gain obtained by optimizing the 2.335M-parameter
backbone. Combining response control and grid control is particularly effective:
$\Phi+\Omega$ uses only 2.15\% of the full trainable state, yet retains 94.6\%
of the full-space accuracy gain and 95.0\% of the full-space logit-error
reduction. Relative to $W$, the same 51.4K-dimensional subspace retains 96.7\%
of the accuracy gain while using roughly $45\times$ fewer trainable parameters.
Thus, quantization correction capacity is not distributed in proportion to
parameter dimension; useful directions are strongly concentrated in structured
low-dimensional subspaces.

\subsection{How Does the Affine Subspace Correct Quantization Error?}
\label{sec:mechanism_empirical}

The subspace comparison shows \emph{where} correction capacity is concentrated.
We next ask \emph{what} the normalization-affine subspace actually corrects.

\paragraph{Local affine error absorption.}
We isolate each of the 53 Conv--BN pairs in MobileNetV2 and optimize only the
corresponding BN affine parameters against the FP response. Across all 53
pairs, the tuned response error is lower
under both W4A4 and W3A3. Mean local error absorption is 17.2\% for W4A4 and
34.0\% for W3A3. This consistency shows that the affine subspace possesses a
broadly distributed ability to absorb structured response distortion rather
than benefiting only a few exceptional layers.

\paragraph{Correction leverage is spatially heterogeneous.}
The block-level intervention compares where that correction is applied inside
17 three-convolution inverted residual blocks. Under W4A4, mean block-output
error absorption increases from 3.65\% with BN1 alone to 11.90\% with
BN1+BN2, 23.85\% with sequential adaptation of all three normalizations, and
27.51\% with BN3 alone. Under W3A3 the corresponding values are 5.85\%,
19.92\%, 50.17\%, and 56.08\%. Terminal-normalization adaptation improves all
17 blocks. We therefore observe that the location where quantization error is
generated need not coincide with the location where it can be corrected most
effectively.

\paragraph{Representation matching is not functional recovery.}
A natural hypothesis is that \sandwichquant{} succeeds by pulling every intermediate
activation back toward its FP counterpart. The depth-wise intervention rejects
this simple explanation. At early depths, \sqpost{} can substantially
\emph{increase} raw FP-relative feature NMSE, yet the same features rapidly
reduce downstream logit error when passed through a frozen FP tail. Thus,
Euclidean restoration of the original intermediate trajectory is not necessary
for functional recovery. Affine adaptation instead reshapes quantized
representations toward states that remain compatible with task-relevant
downstream computation.

\paragraph{Recovered function is encoded in the corrected representation.}
To test whether downstream \sqpost{} layers alone are responsible for recovery,
we perform an interventional feature swap. At each depth, the representation
produced by \sqpost{} is injected into an otherwise unadapted RTN tail. Accuracy
rises progressively from near the RTN baseline to almost the full \sqpost{}
accuracy as the injection point moves deeper. The complementary
RTN-feature-to-\sqpost{}-tail intervention does not rescue
the RTN trajectory at middle and late depths. Together, these interventions
support a distributed correction process: task-relevant function is
progressively encoded in the corrected representation itself rather than being
recovered only by a modified classifier or a few late affine layers.

\paragraph{Response correction has an information-recovery boundary.}
The stronger relative local absorption observed at W3A3 does not mean that
W3A3 is easier to repair. A response--function boundary control compares
global response and functional recovery. Under W4A4, \sqpost{} reduces global logit
NMSE by 74.6\% and recovers 81.9\% of the FP accuracy gap. Under W3A3, the
logit NMSE still falls by 46.8\%, but the accuracy gap recovery is only 0.6\%.
Severe low-bit quantization can therefore remain geometrically correctable in
response magnitude while discriminative information has already been lost.

These results refine the local affine model in
Sec.~\ref{sec:mechanism}: the affine subspace can absorb structured distortion,
but its value is determined by whether those correction directions remain
aligned with downstream task function. In short, representation distance and
functional distance are distinct, and response recovery is not equivalent to
information recovery.

\subsection{Do Affine Directions Remain Useful after QAT?}
\label{sec:qat_results}

The preceding studies start from strongly degraded quantized models. We next
ask whether the same response-control directions remain useful after a model has already undergone QAT. Figure~\ref{fig:qat_diagnostic} provides a
checkpoint-level diagnostic from a common LSQ+ MobileNetV2 checkpoint.

At W4A4, the base checkpoint obtains 68.21\%. Matched extra QAT reaches
68.43\%, whereas $\gamma$-only, $\beta$-only, and joint
$(\gamma,\beta)$ refinement reach 69.65\%, 69.50\%, and 69.83\%,
respectively. Under W2A4, the corresponding joint affine branch improves
63.42\% to 65.86\%. These results show that useful affine response-control
directions can remain after conventional QAT, but they do not by themselves
establish optimizer-independent superiority or the dynamics of a fully
converged full-space objective.

\begin{table*}[t]
\centering
{\small
\setlength{\tabcolsep}{8.0pt}
\begin{tabular}{@{}llccc@{}}
\toprule
\textbf{Model} & \textbf{Method} & \textbf{W/A} &
\textbf{Val. Acc. (\%)} & \textbf{Drop (\%)} \\
\midrule
\multirow{23}{*}{MobileNetV2}
& Full precision & 32/32 & 71.30 & -- \\
\cmidrule(lr){2-5}
& PACT & 4/4 & 64.06 & -7.24 \\
& DSQ & 4/4 & 67.36 & -3.94 \\
& LSQ & 4/4 & 69.01 & -2.29 \\
& LSQ+ & 4/4 & 68.21 & -3.09 \\
& OOQ-Dampen & 4/4 & 66.21 & -5.09 \\
& OOQ-Freeze & 4/4 & 70.02 & -1.28 \\
& StableQAT & 4/4 & 68.24 & -3.06 \\
\rowcolor{oursrow}\cellcolor{white} & \sqpost{} + LSQ+ & 4/4 & 69.83 & -1.47 \\
\rowcolor{oursrow}\cellcolor{white} & \sqpost{} + OOQ-Dampen & 4/4 & 68.07 & -3.23 \\
\rowcolor{oursrow}\cellcolor{white} & \sqpost{} + OOQ-Freeze & 4/4 & \textbf{70.68} & \textbf{-0.62} \\
\rowcolor{oursrow}\cellcolor{white} & \sqpost{} + StableQAT & 4/4 & 69.12 & -2.18 \\
\cmidrule(lr){2-5}
& PACT & 2/4 & 57.90 & -13.40 \\
& DSQ & 2/4 & 62.81 & -8.49 \\
& LSQ & 2/4 & 64.58 & -6.72 \\
& LSQ+ & 2/4 & 63.42 & -7.88 \\
& OOQ-Dampen & 2/4 & 66.23 & -5.07 \\
& OOQ-Freeze & 2/4 & 67.42 & -3.88 \\
& StableQAT & 2/4 & 66.09 & -5.21 \\
\rowcolor{oursrow}\cellcolor{white} & \sqpost{} + LSQ+ & 2/4 & 65.86 & -5.44 \\
\rowcolor{oursrow}\cellcolor{white} & \sqpost{} + OOQ-Dampen & 2/4 & 68.25 & -3.05 \\
\rowcolor{oursrow}\cellcolor{white} & \sqpost{} + OOQ-Freeze & 2/4 & 62.81 & -8.49 \\
\rowcolor{oursrow}\cellcolor{white} & \sqpost{} + StableQAT & 2/4 & \textbf{68.50} & \textbf{-2.80} \\
\bottomrule
\end{tabular}
}
\caption{
Post-QAT adaptation on CIFAR-100 with MobileNetV2.
Drop denotes the signed accuracy difference from the 71.30\% FP32 baseline.
}
\label{tab:cifar100_qat_diagnostic}
\end{table*}

Table~\ref{tab:cifar100_qat_diagnostic} broadens this checkpoint-level test
across several QAT methods. At W4A4, \sandwichquant{}
improves LSQ+, OOQ-Dampen, OOQ-Freeze, and StableQAT, with the strongest result
of $70.68\%$ obtained on OOQ-Freeze. At W2A4, \sandwichquant{} improves LSQ+,
OOQ-Dampen, and StableQAT, and \sandwichquant{}+StableQAT reaches $68.50\%$. However,
\sandwichquant{} degrades the W2A4 OOQ-Freeze checkpoint from $67.42\%$ to $62.81\%$.
Thus, the affine subspace remains useful after QAT, but its benefit is not
unconditional: when a checkpoint already encodes a highly specialized
low-bit parameterization, an additional affine update may interfere with the
learned quantization solution. This negative case motivates backend- and
checkpoint-aligned adaptation rather than a universal post-hoc claim.

\subsection{Extreme PTQ on ImageNet}
\label{sec:imagenet_main}

Table~\ref{tab:main_imagenet_mnv2} shows that the original MobileNetV2
parameterization is extremely fragile under full weight-and-activation
quantization: RTN falls from $71.87\%$ to $0.33\%$ at W4A4 and to $0.11\%$ at
W3A4. Updating only normalization affine parameters recovers W4A4 accuracy to
$66.11\%$, exceeding QDrop by 5.40 points without updating convolutional or
linear weights. At W3A4, \sandwichquant{} alone and QDrop are similar, while
\sandwichquant{}+QDrop reaches the best accuracy of $53.45\%$. These results support two
conclusions. First, the pretrained response parameterization is a major source
of extreme low-bit fragility. Second, \sandwichquant{} and reconstruction are complementary
only when their fake-quantization graphs and operating bit-widths are aligned;
the combination is not uniformly superior in every setting.

\subsection{Cross-Task and Cross-Architecture Generalization}
\label{sec:cross_task}

The Cityscapes results in Table~\ref{tab:cityscapes_unet} show that the
normalization affine subspace is not specific to image classification. Direct
RTN collapses to $3.27\%$ mIoU at W4A4, whereas \sandwichquant{} restores $67.21\%$, only
1.31 points below the FP32 model and 0.76 points above QDrop. Under the more
aggressive W3A4 setting, \sandwichquant{}+QDrop obtains $61.36\%$, improving QDrop by 3.15
points. The same pattern as ImageNet emerges: \sandwichquant{} alone is particularly strong
at W4A4, while its combination with reconstruction becomes more useful as the
bit-width decreases.

\begin{table*}[t]
\centering
\begin{minipage}[t]{0.485\textwidth}
\vspace{0pt}
\centering
{\footnotesize
\setlength{\tabcolsep}{2.0pt}
\resizebox{\linewidth}{!}{%
\begin{tabular}{@{}llccc@{}}
\toprule
\multirow{2}{*}{\textbf{Model}}
& \multirow{2}{*}{\textbf{Method}}
& \multirow{2}{*}{\textbf{FP32}}
& \multicolumn{2}{c}{\textbf{Quant. Acc.} $\uparrow$} \\
\cmidrule(lr){4-5}
& & & \textbf{W4A4} & \textbf{W3A4} \\
\midrule
\multirow{4}{*}{MobileNetV2}
& RTN & & 0.33 & 0.11 \\
& QDrop & & 60.71 & 51.75 \\
\rowcolor{oursrow}\cellcolor{white} & \sqpost{} & & \textbf{66.11} & 51.07 \\
\rowcolor{oursrow}\cellcolor{white} & \sqpre{} + QDrop & & 62.91 & \textbf{53.45} \\
\bottomrule
\end{tabular}}
}
\captionof{table}{ImageNet-1K results with MobileNetV2. Results are top-1
accuracy (\%).}
\label{tab:main_imagenet_mnv2}
\end{minipage}
\hfill
\begin{minipage}[t]{0.485\textwidth}
\vspace{0pt}
\centering
{\footnotesize
\setlength{\tabcolsep}{2.0pt}
\resizebox{\linewidth}{!}{%
\begin{tabular}{@{}llccc@{}}
\toprule
\multirow{2}{*}{\textbf{Arch.}}
& \multirow{2}{*}{\textbf{Method}}
& \multirow{2}{*}{\textbf{FP32}}
& \multicolumn{2}{c}{\textbf{Quantized mIoU} $\uparrow$} \\
\cmidrule(lr){4-5}
& & & \textbf{W4A4} & \textbf{W3A4} \\
\midrule
\multirow{4}{*}{U-Net}
& RTN & \multirow{4}{*}{68.52} & 3.27 & 1.24 \\
& QDrop & & 66.45 & 58.21 \\
\rowcolor{oursrow}\cellcolor{white} & \sqpost{} & & \textbf{67.21} & 59.07 \\
\rowcolor{oursrow}\cellcolor{white} & \sqpre{} + QDrop & & 66.78 & \textbf{61.36} \\
\bottomrule
\end{tabular}}
}
\captionof{table}{Cityscapes semantic segmentation with U-Net. Results are
mIoU (\%).}
\label{tab:cityscapes_unet}
\end{minipage}
\end{table*}

\begin{table*}[t]
\centering
\caption{\textbf{Order ablation of normalization-affine adaptation.}
We compare post-quantization correction
($\mathrm{PTQ}\!\rightarrow\!\Phi_{\mathrm{post}}$),
quantization preconditioning
($\Phi_{\mathrm{pre}}\!\rightarrow\!\mathrm{PTQ}$),
and their composition.
\textsc{Sandwich-Quant} applies independently calibrated normalization-affine
adaptation before and after PTQ:
$\Phi_{\mathrm{pre}}\!\rightarrow\!\mathrm{PTQ}
 \!\rightarrow\!\Phi_{\mathrm{post}}$.
Wiki2 and C4 report perplexity ($\downarrow$); Avg6 is the unweighted
mean accuracy over six zero-shot tasks ($\uparrow$).
All affine stages use the same per-stage calibration data, objective, update
budget, and seed; composed variants contain the stages indicated by the row.}
\label{tab:order_ablation}

{\small
\setlength{\tabcolsep}{3.6pt}
\renewcommand{\arraystretch}{1.08}
\resizebox{\textwidth}{!}{%
\begin{tabular}{@{}l|rrr|rrr|rrr|rrr@{}}
\toprule
& \multicolumn{6}{c|}{\textbf{Llama3-8B}}
& \multicolumn{6}{c}{\textbf{Qwen3-8B}} \\
\cmidrule(lr){2-7}
\cmidrule(lr){8-13}

\textbf{Order}
& \multicolumn{3}{c|}{\textbf{W3A16}}
& \multicolumn{3}{c|}{\textbf{W2A4KV4}}
& \multicolumn{3}{c|}{\textbf{W3A16}}
& \multicolumn{3}{c}{\textbf{W2A4KV4}} \\
\cmidrule(lr){2-4}
\cmidrule(lr){5-7}
\cmidrule(lr){8-10}
\cmidrule(lr){11-13}

& \textbf{Wiki2$\downarrow$} & \textbf{C4$\downarrow$}
& \textbf{Avg6$\uparrow$}
& \textbf{Wiki2$\downarrow$} & \textbf{C4$\downarrow$}
& \textbf{Avg6$\uparrow$}
& \textbf{Wiki2$\downarrow$} & \textbf{C4$\downarrow$}
& \textbf{Avg6$\uparrow$}
& \textbf{Wiki2$\downarrow$} & \textbf{C4$\downarrow$}
& \textbf{Avg6$\uparrow$} \\
\midrule

$\mathrm{PTQ}$
& 8.33 & 13.10 & 68.83
& 18.07 & 50.56 & 46.81
& 11.29 & 17.09 & 71.87
& 18.71 & 38.27 & 49.62 \\

$\mathrm{PTQ}\rightarrow\Phi_{\mathrm{post}}$
& 7.89 & 12.21 & \textbf{70.08}
& 12.83 & 30.35 & 52.49
& 10.60 & 16.38 & 72.17
& 12.73 & 26.80 & 56.47 \\

$\Phi_{\mathrm{pre}}\rightarrow\mathrm{PTQ}$
& 8.28 & 12.87 & 68.89
& 13.93 & 37.96 & 48.72
& 10.74 & 16.42 & 72.31
& 12.99 & 27.99 & 55.28 \\

\rowcolor{oursrow}
\textbf{\sandwichquant}
& \textbf{7.87} & \textbf{12.13} & 70.05
& \textbf{12.31} & \textbf{28.86} & \textbf{53.44}
& \textbf{10.51} & \textbf{16.15} & \textbf{72.34}
& \textbf{12.03} & \textbf{25.37} & \textbf{58.05} \\

\bottomrule
\end{tabular}%
}}
\end{table*}

}

\subsection{Cross-Domain and Checkpoint Controls}
\label{sec:cross_domain_controls}
Before turning to LLMs, we test whether affine response control is specific to
one architecture or training regime.  Table~\ref{tab:cifar100_qat_diagnostic}
shows that a short post-QAT affine stage improves checkpoints produced by
several QAT algorithms at both W4A4 and W2A4.  The W2A4 OOQ-Freeze exception is
also informative: affine adaptation is a structured correction mechanism, not
an unconditional post-hoc improvement.

\begin{table*}[t]
\centering
{\small
\setlength{\tabcolsep}{8.0pt}
\begin{tabular}{@{}llccc@{}}
\toprule
\textbf{Model} & \textbf{Method} & \textbf{W/A} &
\textbf{Val. Acc. (\%)} & \textbf{Drop (\%)} \\
\midrule

& Full precision & 32/32 & 71.30 & -- \\
\cmidrule(lr){2-5}
& PACT & 4/4 & 64.06 & -7.24 \\
& DSQ & 4/4 & 67.36 & -3.94 \\
& LSQ & 4/4 & 69.01 & -2.29 \\
& LSQ+ & 4/4 & 68.21 & -3.09 \\
\rowcolor{oursrow}\cellcolor{white} & \sqpost{} + LSQ+ & 4/4 & 69.83 & -1.47 \\
& OOQ-Dampen & 4/4 & 66.21 & -5.09 \\
\rowcolor{oursrow}\cellcolor{white} & \sqpost{} + OOQ-Dampen & 4/4 & 68.07 & -3.23 \\
& OOQ-Freeze & 4/4 & 70.02 & -1.28 \\
\rowcolor{oursrow}\cellcolor{white} & \sqpost{} + OOQ-Freeze & 4/4 & \textbf{70.68} & \textbf{-0.62} \\
& StableQAT & 4/4 & 68.24 & -3.06 \\
\rowcolor{oursrow}\cellcolor{white} & \sqpost{} + StableQAT & 4/4 & 69.12 & -2.18 \\
\cmidrule(lr){2-5}
& PACT & 2/4 & 57.90 & -13.40 \\
& DSQ & 2/4 & 62.81 & -8.49 \\
& LSQ & 2/4 & 64.58 & -6.72 \\
& LSQ+ & 2/4 & 63.42 & -7.88 \\
\rowcolor{oursrow}\cellcolor{white} & \sqpost{} + LSQ+ & 2/4 & 65.86 & -5.44 \\
& OOQ-Dampen & 2/4 & 66.23 & -5.07 \\
\rowcolor{oursrow}\cellcolor{white} & \sqpost{} + OOQ-Dampen & 2/4 & 68.25 & -3.05 \\
& OOQ-Freeze & 2/4 & 67.42 & -3.88 \\
\rowcolor{oursrow}\cellcolor{white} & \sqpost{} + OOQ-Freeze & 2/4 & 62.81 & -8.49 \\
& StableQAT & 2/4 & 66.09 & -5.21 \\
\rowcolor{oursrow}\cellcolor{white} \multirow{-23}{*}{MobileNetV2} & \sqpost{} + StableQAT & 2/4 & \textbf{68.50} & \textbf{-2.80} \\
\bottomrule
\end{tabular}
}
\caption{
One-sided post-QAT affine adaptation on CIFAR-100 with MobileNetV2.
Highlighted rows apply only \sqpost{} after the named QAT checkpoint; they are
component controls rather than the complete two-stage \sandwichquant{} pipeline.
Drop denotes the signed accuracy difference from the 71.30\% FP32 baseline.
}
\label{tab:cifar100_qat_diagnostic}
\end{table*}

The same phenomenon transfers across tasks and architectures
(Table~\ref{tab:main_imagenet_mnv2} and Table~\ref{tab:cityscapes_unet}).
At W4A4, affine-only correction restores MobileNetV2 from 0.33\% to 66.11\%
top-1 and U-Net from 3.27\% to 67.21\% mIoU.  At W3A4, combining affine
preconditioning with QDrop is strongest, indicating growing complementarity
as quantization becomes more destructive.

\begin{table*}[t]
\centering
\begin{minipage}[t]{0.49\textwidth}
\vspace{0pt}\centering
{\scriptsize\setlength{\tabcolsep}{1.8pt}
\renewcommand{\arraystretch}{1.05}
\resizebox{\linewidth}{!}{%
\begin{tabular}{@{}llccc@{}}
\toprule
\multirow{2}{*}{\textbf{Model}} & \multirow{2}{*}{\textbf{Method}} &
\multirow{2}{*}{\textbf{FP32}} &
\multicolumn{2}{c}{\textbf{Quant. Acc.} $\uparrow$} \\
\cmidrule(lr){4-5}
& & & \textbf{W4A4} & \textbf{W3A4} \\
\midrule
& RTN & & 0.33 & 0.11 \\
\rowcolor{oursrow}\cellcolor{white}
& \textsc{SQ-Post}
& & \textbf{66.11} & 51.07 \\
& QDrop & & 60.71 & 51.75 \\
\rowcolor{oursrow}\cellcolor{white}\multirow{-4}{*}{MobileNetV2}
& \textsc{SQ-Pre} + QDrop
& \cellcolor{white}\multirow{-4}{*}{71.87} & 62.91 & \textbf{53.45} \\
\bottomrule
\end{tabular}}}
\captionof{table}{ImageNet-1K top-1 accuracy (\%) with MobileNetV2.
\textsc{SQ-Post} and \textsc{SQ-Pre} abbreviate the corresponding one-sided
components of \sandwichquant{}.}
\label{tab:main_imagenet_mnv2}
\end{minipage}
\hfill
\begin{minipage}[t]{0.49\textwidth}
\vspace{0pt}\centering
{\scriptsize\setlength{\tabcolsep}{1.8pt}
\renewcommand{\arraystretch}{1.05}
\resizebox{\linewidth}{!}{%
\begin{tabular}{@{}llccc@{}}
\toprule
\multirow{2}{*}{\textbf{Model}} & \multirow{2}{*}{\textbf{Method}} &
\multirow{2}{*}{\textbf{FP32}} &
\multicolumn{2}{c}{\textbf{Quantized mIoU} $\uparrow$} \\
\cmidrule(lr){4-5}
& & & \textbf{W4A4} & \textbf{W3A4} \\
\midrule
& RTN & & 3.27 & 1.24 \\
\rowcolor{oursrow}\cellcolor{white}
& \textsc{SQ-Post}
& & \textbf{67.21} & 59.07 \\
& QDrop & & 66.45 & 58.21 \\
\rowcolor{oursrow}\cellcolor{white}\multirow{-4}{*}{U-Net}
& \textsc{SQ-Pre} + QDrop
& \cellcolor{white}\multirow{-4}{*}{68.52} & 66.78 & \textbf{61.36} \\
\bottomrule
\end{tabular}}}
\captionof{table}{Cityscapes mIoU (\%) with U-Net.
\textsc{SQ-Post} and \textsc{SQ-Pre} abbreviate the corresponding one-sided
components of \sandwichquant{}.}
\label{tab:cityscapes_unet}
\end{minipage}
\end{table*}

Finally, Table~\ref{tab:order_ablation} isolates the two sides of the LLM
pipeline.  Post-PTQ correction is the more stable single stage, whereas the
full composition is strongest in three of four settings and yields its largest
gain under W2A4KV4.  We therefore use the complete composition as
\sandwichquant{} in the headline tables below.

\begin{table*}[t]
\centering
\caption{\textbf{Order ablation of normalization-affine adaptation.}
$\Phi_{\mathrm{pre}}$ preconditions PTQ, whereas $\Phi_{\mathrm{post}}$
corrects the frozen quantized graph.  Wiki2/C4 are perplexities ($\downarrow$)
and Avg6 is mean zero-shot accuracy ($\uparrow$). Each affine stage uses the
same per-stage budget; \sandwichquant{} contains two affine stages and rebuilds
the PTQ graph from the original dense checkpoint between them.}
\label{tab:order_ablation}
{\small
\setlength{\tabcolsep}{3.1pt}
\renewcommand{\arraystretch}{0.98}
\resizebox{\textwidth}{!}{%
\begin{tabular}{@{}l|rrr|rrr|rrr|rrr@{}}
\toprule
& \multicolumn{6}{c|}{\textbf{Llama3-8B}}
& \multicolumn{6}{c}{\textbf{Qwen3-8B}} \\
\cmidrule(lr){2-7}\cmidrule(lr){8-13}
\textbf{Order}
& \multicolumn{3}{c|}{\textbf{W3A16}}
& \multicolumn{3}{c|}{\textbf{W2A4KV4}}
& \multicolumn{3}{c|}{\textbf{W3A16}}
& \multicolumn{3}{c}{\textbf{W2A4KV4}} \\
\cmidrule(lr){2-4}\cmidrule(lr){5-7}\cmidrule(lr){8-10}\cmidrule(lr){11-13}
& \textbf{Wiki2$\downarrow$} & \textbf{C4$\downarrow$} & \textbf{Avg6$\uparrow$}
& \textbf{Wiki2$\downarrow$} & \textbf{C4$\downarrow$} & \textbf{Avg6$\uparrow$}
& \textbf{Wiki2$\downarrow$} & \textbf{C4$\downarrow$} & \textbf{Avg6$\uparrow$}
& \textbf{Wiki2$\downarrow$} & \textbf{C4$\downarrow$} & \textbf{Avg6$\uparrow$} \\
\midrule
$\mathrm{PTQ}$
& 8.33 & 13.10 & 68.83 & 18.07 & 50.56 & 46.81
& 11.29 & 17.09 & 71.87 & 18.71 & 38.27 & 49.62 \\
$\mathrm{PTQ}\!\rightarrow\!\Phi_{\mathrm{post}}$
& 7.89 & 12.21 & \textbf{70.08} & 12.83 & 30.35 & 52.49
& 10.60 & 16.38 & 72.17 & 12.73 & 26.80 & 56.47 \\
$\Phi_{\mathrm{pre}}\!\rightarrow\!\mathrm{PTQ}$
& 8.28 & 12.87 & 68.89 & 13.93 & 37.96 & 48.72
& 10.74 & 16.42 & 72.31 & 12.99 & 27.99 & 55.28 \\
\rowcolor{oursrow}
\textbf{\sandwichquant}
& \textbf{7.87} & \textbf{12.13} & 70.05
& \textbf{12.31} & \textbf{28.86} & \textbf{53.44}
& \textbf{10.51} & \textbf{16.15} & \textbf{72.34}
& \textbf{12.03} & \textbf{25.37} & \textbf{58.05} \\
\bottomrule
\end{tabular}}}
\end{table*}

\subsection{Low-Bit Quantization of Large Language Models}
\label{sec:llm_weight_only}

\begin{table*}[t]
\centering
\caption{\textbf{Weight-only W3A16 results (group size 128).}
Highlighted rows apply the complete \sandwichquant{} pipeline to the backend
named in parentheses. Avg. is the unweighted mean over the six zero-shot tasks.
$\mathrm{ResComp}^{\ast}$ uses GPTAQ as its base quantizer. All variants use the
same 128 C4 sequences (length 2,048; seed 0); full settings are in
Appendix~\ref{app:llm_protocol}.}
\label{tab:llm_w3_main}

\resizebox{\textwidth}{!}{%
\begin{tabular}{c|l|rr|rrrrrr|c}
\hline\hline
\multicolumn{1}{c|}{\multirow{2}{*}{\textbf{Model}}}
& \multicolumn{1}{c|}{\multirow{2}{*}{\textbf{Method}}}
& \multicolumn{2}{c|}{\textbf{Perplexity $\downarrow$}}
& \multicolumn{6}{c|}{\textbf{Zero-shot Accuracy (\%) $\uparrow$}}
& \multirow{2}{*}{\textbf{Avg. $\uparrow$}} \\
\cline{3-10}
& & \textbf{Wiki2} & \textbf{C4}
& \textbf{PIQA} & \textbf{ARC-E} & \textbf{ARC-C}
& \textbf{HS} & \textbf{WG} & \textbf{BoolQ} & \\
\hline

& FP16 & 5.47 & 7.27 & 78.9 & 74.6 & 46.1 & 75.9 & 69.2 & 77.7 & 70.4 \\
\cline{2-11}
& RTN & 8.22 & 11.53  & 76.3  & 66.5  & 40.6 & 68.7 & 65.8 & 67.5 & 64.2 \\
\rowcolor{oursrow}
\cellcolor{white} & \sandwichquant{} (RTN) & 7.40  &  10.29 & 77.4  & 68.9  & 41.9 & 71.0 & 66.2 & 70.9 & 66.1 \\
& AWQ & 6.79  & 8.93  & 77.1  & 69.9  & 41.8 & 71.2 & 67.7 & 71.5 & 66.3 \\
\rowcolor{oursrow}
\cellcolor{white} & \sandwichquant{} (AWQ) &  6.47 & 8.55  &  77.5 & \textbf{70.8}  & \textbf{43.1} & 72.4 & 67.3 & 72.9 & 67.3 \\
& GPTQ & 6.75 & 13.72 & 76.8 & 66.5 & 39.7 & 68.3 & 67.5 & 68.9 & 64.6 \\
\rowcolor{oursrow}
\cellcolor{white} & \sandwichquant{} (GPTQ) & 6.28 & 8.31 & 77.3 & 66.3 & 41.6 & 72.1 & \textbf{67.8} & 69.4 & 65.8 \\
& GPTAQ &  6.81 & 8.40  & 77.8  & 69.2  & 40.7 & 71.9 & 67.6 & 71.5 & 66.5 \\
\rowcolor{oursrow}
\cellcolor{white} & \sandwichquant{} (GPTAQ) & 6.27  & 8.16  & \textbf{78.3}  & 70.5  & 40.8 & 72.3 & 66.7 & 71.3 & 66.7 \\
& $\mathrm{ResComp}^{\ast}$ & 6.25  & 8.19  & 77.4  & 69.5  & 41.5 & 72.3 & 67.4 & 71.5 & 66.6 \\
\rowcolor{oursrow}
\cellcolor{white} \multirow{-11}{*}{Llama2-7B} & \sandwichquant{} ($\mathrm{ResComp}^{\ast}$) & \textbf{6.18}  &  \textbf{8.12} & 77.5  &  69.6 & 42.8 & \textbf{72.9} & 67.1 & \textbf{73.0} & \textbf{67.2} \\
\hline

& FP16 & 6.14 & 9.45 & 80.9 & 77.7 & 53.2 & 79.2 & 72.9 & 81.2 & 74.2 \\
\cline{2-11}
& RTN & 29.21 & 43.1 & 68.5 & 50.1 & 35.4 & 53.5 & 58.8 & 63.7 & 55.0 \\
\rowcolor{oursrow}
\cellcolor{white} & \sandwichquant{} (RTN) & 15.1 & 22.0 & 73.2 & 57.5 & 35.4 & 63.2 & 63.3 & 67.0 & 59.9 \\
& AWQ & 9.53 & 14.74 & 76.1 & 69.2 & 42.2 & 71.4 & 69.0 & 78.2 & 67.7 \\
\rowcolor{oursrow}
\cellcolor{white} & \sandwichquant{} (AWQ) & 8.97 & 14.09 & 77.8 & 71.0 & 44.5 & 71.6 & 70.1 & 76.3 & 68.6 \\
& GPTQ & 8.33 & 13.11 & 77.5 & 70.6 & 43.7 & 71.9 & 71.4 & 77.9 & 68.8 \\
\rowcolor{oursrow}
\cellcolor{white} & \sandwichquant{} (GPTQ) & 7.87 & 12.13 & 78.1 & 72.3 & 44.2 & 74.2 & 72.3 & \textbf{79.2} & 70.1 \\
& GPTAQ &  8.13 & 12.77  & 77.3  & 69.3  & 43.6 & 61.9 & 72.0 & 77.7 & 67.0 \\
\rowcolor{oursrow}
\cellcolor{white} & \sandwichquant{} (GPTAQ) & 7.85  & 12.28  & 77.3  &  69.7 & 44.6 & 67.8 & \textbf{72.5} & 78.1 & 68.3 \\
& $\mathrm{ResComp}^{\ast}$ & 7.77 & 12.25 & 77.7 & 73.8 & 45.7 & 74.6 & 72.3 & 79.1 & 70.5 \\
\rowcolor{oursrow}
\cellcolor{white} \multirow{-11}{*}{Llama3-8B} & \sandwichquant{} ($\mathrm{ResComp}^{\ast}$) & \textbf{7.62} & \textbf{12.08} & \textbf{79.2} & \textbf{75.2} & \textbf{48.1} & \textbf{75.3} & 72.4 & 77.2 & \textbf{71.2} \\
\hline

& FP16 & 9.72 & 15.40 & 77.8 & 80.9 & 56.7 & 74.9 & 67.8 & 86.6 &  74.1\\
\cline{2-11}
& RTN & 23.54 & 33.86 & 68.2 & 58.7 & 35.3 & 54.6 & 54.9 & 63.9 & 55.9 \\
\rowcolor{oursrow}
\cellcolor{white} & \sandwichquant{} (RTN) & 16.56 & 23.64 & 73.9 & 68.8 & 43.5 & 63.7 & 63.2 & 76.6 & 65.0 \\
& AWQ & 12.11 & 18.49 & 74.3 & 71.3 & 45.8 & 67.8 & 63.3 & 82.5 & 67.5 \\
\rowcolor{oursrow}
\cellcolor{white} & \sandwichquant{} (AWQ) & 11.50 & 17.44 & 74.8 & 71.6 & 46.9 & 68.5 & 64.8 & 82.9 & 68.3 \\
& GPTQ & 11.29 & 17.09 & 76.9 & 76.8 & 50.9 & 71.6 & \textbf{69.2} & \textbf{85.8} & 71.9 \\
\rowcolor{oursrow}
\cellcolor{white} & \sandwichquant{} (GPTQ) & \textbf{10.51} & \textbf{16.15} & 76.8 & 76.8 & 52.1 & \textbf{72.6} & \textbf{69.4} & \textbf{85.8} & 72.3 \\
& GPTAQ & 11.37  & 17.13  & 75.7  & 72.2  & 48.7 & 70.9 & 67.9 & 85.7 & 70.2 \\
\rowcolor{oursrow}
\cellcolor{white} & \sandwichquant{} (GPTAQ) & 10.68  & 16.42  & 76.3  & 72.1  & 48.9 & 71.5 & 68.5 & 85.0 & 70.4 \\
& $\mathrm{ResComp}^{\ast}$ & 11.66  & 17.45  & 76.6  & 77.6  & 53.2 & 70.7 & 67.3 & 84.7 &  71.7 \\
\rowcolor{oursrow}
\cellcolor{white} \multirow{-11}{*}{Qwen3-8B} & \sandwichquant{} ($\mathrm{ResComp}^{\ast}$) &  \textbf{10.51} &  16.15 & \textbf{77.5}  &  \textbf{78.3} & \textbf{54.6} & 71.3 & 67.9 & 84.7 & \textbf{72.4} \\
\hline\hline
\end{tabular}}
\end{table*}

Table~\ref{tab:llm_w3_main} shows that the complete two-sided pipeline improves
every tested backend and model family. The gains are largest for fragile RTN
and remain measurable after compensated GPTAQ/ResComp, indicating that the
affine stages address residual response geometry rather than merely replacing
the base quantizer. One-sided \sqpost{} counterparts and diagnostic baselines are
reported separately in Appendix~\ref{app:llm_details}.

\begin{table*}[t]
\centering
\caption{\textbf{Joint W2A4KV4 results (group size 128).}
Highlighted rows apply the complete \sandwichquant{} pipeline to the indicated
QuaRot backend. Avg. is the unweighted mean over six zero-shot tasks.
$\mathrm{ResComp}^{\ast}$ uses GPTAQ as its base quantizer. All variants use
the same 128 WikiText-2 sequences (length 2,048; seed 0); full settings are in
Appendix~\ref{app:llm_protocol}.}
\label{tab:llm_w2a4kv4_main}

\resizebox{\textwidth}{!}{%
\begin{tabular}{c|l|rr|rrrrrr|c}
\hline\hline
\multicolumn{1}{c|}{\multirow{2}{*}{\textbf{Model}}}
& \multicolumn{1}{c|}{\multirow{2}{*}{\textbf{Method}}}
& \multicolumn{2}{c|}{\textbf{Perplexity $\downarrow$}}
& \multicolumn{6}{c|}{\textbf{Zero-shot Accuracy (\%) $\uparrow$}}
& \multirow{2}{*}{\textbf{Avg. $\uparrow$}} \\
\cline{3-10}
& & \textbf{Wiki2} & \textbf{C4}
& \textbf{PIQA} & \textbf{ARC-E} & \textbf{ARC-C}
& \textbf{HS} & \textbf{WG} & \textbf{BoolQ} & \\
\hline

& FP16 & 5.47 & 7.27 & 78.9 & 74.6 & 46.1 & 75.9 & 69.2 & 77.7 & 70.4 \\
\cline{2-11}
& QuaRot + GPTAQ & 11.7 & 24.8 & 62.3 & 45.6 & 25.6 & 41.2 & 54.0 & 62.4 & 48.5 \\
\rowcolor{oursrow}
\cellcolor{white} & \sandwichquant{} (QuaRot+GPTAQ) & 8.2 & \textbf{14.0} & 66.8 & 50.0 & \textbf{31.6} & \textbf{54.9} & 58.2 & 64.5 & 54.3 \\
& QuaRot + $\mathrm{ResComp}^{\ast}$ & 11.5 & 23.6 & 63.6 & 46.4 & 24.9 & 40.8 & 56.0 & 61.9 & 48.9 \\
\rowcolor{oursrow}
\cellcolor{white}\multirow{-5}{*}{Llama2-7B} & \sandwichquant{} (QuaRot+$\mathrm{ResComp}^{\ast}$) & \textbf{8.1} & 14.2 & \textbf{68.1} & \textbf{52.2} & 30.7 & 54.8 & \textbf{59.1} & \textbf{65.8} & \textbf{55.1}  \\
\hline

& FP16 & 6.14 & 9.45 & 80.9 & 77.7 & 53.2 & 79.2 & 72.9 & 81.2 & 74.2 \\
\cline{2-11}
& QuaRot + GPTAQ & 23.0 & 62.8 & 55.8 & 37.2 & 22.3 & 36.5 & 50.4 & 59.3 & 43.6 \\
\rowcolor{oursrow}
\cellcolor{white} & \sandwichquant{} (QuaRot+GPTAQ) & 12.2 & 28.7 & 63.9 & 49.8 & 29.3 & 51.0 & 57.2 & 64.9 & 52.7 \\
& QuaRot + $\mathrm{ResComp}^{\ast}$ & 22.1 & 61.5 & 56.0 & 38.5 & 24.6 & 35.8 & 53.5 & 60.9 & 44.9 \\
\rowcolor{oursrow}
\cellcolor{white} \multirow{-5}{*}{Llama3-8B} & \sandwichquant{} (QuaRot+$\mathrm{ResComp}^{\ast}$) & 12.2 & 29.1 & 65.6 & 49.7 & 29.8 & 50.7 & 58.5 & 66.3 & 53.4 \\
\hline

& FP16 & 9.72 & 15.40 & 77.8 & 80.9 & 56.7 & 74.9 & 67.8 & 86.6 &  74.1\\
\cline{2-11}
& QuaRot + GPTAQ & 28.2 & 64.1 & 57.9 & 39.5 & 23.4 & 34.7 & 51.7 & 61.6 & 44.8 \\
\rowcolor{oursrow}
\cellcolor{white} & \sandwichquant{} (QuaRot+GPTAQ) & 14.6 & 32.8 & 66.8 & 52.4 & 32.5 & 47.9 & 59.1 & 66.9 & 54.3 \\
& QuaRot + $\mathrm{ResComp}^{\ast}$ & 18.7 & 38.3 & 61.6 & 44.4 & 28.6 & 43.7 & 54.8 & 64.4 & 49.6 \\
\rowcolor{oursrow}
\cellcolor{white} \multirow{-5}{*}{Qwen3-8B} & \sandwichquant{} (QuaRot+$\mathrm{ResComp}^{\ast}$) & 12.0 & 25.4 & 69.1 & 56.8 & 37.1 & 53.6 & 62.1 & 70.1 & 58.1 \\

\hline\hline
\end{tabular}}
\end{table*}

Joint activation and KV-cache quantization amplifies the advantage of using
both affine stages. Table~\ref{tab:llm_w2a4kv4_main} shows especially large
perplexity and Avg. gains on Llama3-8B and Qwen3-8B, while
Table~\ref{tab:order_ablation} attributes these gains to complementary
preconditioning and frozen-graph correction rather than a longer post-tuning
run alone.

\long\def\deferredcontrols{
\subsection{Additional Controls and Practical Boundaries}
\label{sec:why_sq}

\paragraph{Matched-size controls: structure, not sparsity alone.}
Figure~\ref{fig:matched_subspace_qwen3} and
Table~\ref{tab:parameter_subspace} ask whether arbitrary weight coordinates of
the same size work equally well.  On Qwen3-8B, $\Phi$ slightly outperforms even
gradient-selected weights under an identical 299,008-parameter budget.  On
ImageNet, its 34.2K coordinates reach 66.11\% W4A4 accuracy, versus 18.35\% for
the strongest equal-size weight control.  Low dimensionality alone therefore
does not explain the effect; the normalization-affine structure matters.

\paragraph{Knowledge distillation and backend alignment.}
A separate control separates teacher guidance from graph alignment.
Removing KD reduces W4A4 accuracy from $66.11\%$ to $61.50\%$, indicating that
label supervision alone does not preserve the teacher decision structure as
effectively. More importantly, transferring a \sandwichquant{} checkpoint to a mismatched
fake-quantized graph yields only $48.72\%$. The 17.39-point gap to matched \sandwichquant{}
confirms that \sandwichquant{} learns backend-specific compensation rather than a generic
channel rescaling.

\paragraph{Parameter efficiency is not data efficiency.}
\sandwichquant{} maintains optimizer states for only 34.2K parameters and requires
$1.00\times$ normalized time, compared with $1.83\times$ for full tuning.
However, Table~\ref{tab:data_dependence} shows that the current ImageNet
implementation is strongly dependent on tuning-data coverage. Using 5K or 10K
images fails to recover the collapsed model, whereas full-data tuning reaches
$66.11\%$. \sandwichquant{} is therefore parameter-efficient but, in its present form, not
a calibration-only or few-shot method. This distinction is important when
comparing it with conventional PTQ methods that use only a small calibration
set.

Overall, the controls show that quantization-relevant correction leverage is
highly non-uniform across parameter coordinates, with $\Phi$ occupying a
particularly favorable capacity--dimension operating point. They also expose
clear boundaries: the adaptation is
backend-specific, can conflict with some QAT solutions, depends on sufficient
tuning-data coverage in the current ImageNet setting, and cannot recover task
information destroyed by severe clipping and rounding.

}

\section{Discussion and Limitations}

\paragraph{What the evidence supports.}
For the same small parameter budget, normalization-affine coordinates provide
unusually effective correction directions, whose benefit emerges primarily
after residual errors propagate. Figures~\ref{fig:matched_subspace_qwen3} and
\ref{fig:mechanism_joint} do not establish exact projection or uniform local
reconstruction; they show that task-aware damage can fall even when an isolated
local discrepancy changes little or increases.

\paragraph{Scope and cost.}
The pre-stage changes the model presented to PTQ, whereas the post-stage adapts
to the realized frozen graph. Their composition is most useful under aggressive
joint quantization (Table~\ref{tab:order_ablation}). It adds no inference
operator, but requires two short affine optimizations, a second offline PTQ
pass, and a trainable target graph during calibration. Affine states are
backend-, bit-width-, and calibration-specific, and cannot restore information
destroyed by severe clipping or rounding. Evidence is limited to dense 7B--8B
LLMs and one deterministic seed; larger and MoE models, deployment latency, and
multi-seed confidence intervals remain future work.

\section{Conclusion}

We cast low-bit quantization correction as a parameter-subspace problem and
identify normalization-affine coordinates as unusually high-leverage response
controls. This motivates \sandwichquant{}, which independently adapts the same
native affine subspace before and after PTQ. Across three LLM families and
weight-only and joint weight--activation--KV regimes, the two-stage construction
consistently improves perplexity and mean downstream accuracy without adding an
inference-time operator. Its limits are equally clear: it corrects structured,
reachable response distortion, not information irreversibly lost to clipping
or rounding.

\section*{Large Language Model (LLM) Usage}
In this paper, LLMs assisted with language polishing and limited manuscript-production
support. All scientific decisions, experiments, result verification, and
conclusions were performed or approved by the authors, who take full
responsibility for the manuscript.

\bibliographystyle{plainnat}

\bibliography{ref}

\appendix
\section{Reproducibility and Complete Experimental Protocol}
\label{app:llm_protocol}

\paragraph{Two-stage construction.}
For every backend $\mathcal B$, we first run PTQ on a fixed calibration tensor
and optimize only normalization-affine parameters on the completed quantized
graph. The resulting affine state is used as $\Phi_{\mathrm{pre}}$: only these
affine tensors are transferred to the original dense checkpoint; all weights,
quantizer variables, and optimizer state from the first run are discarded.
We then rerun $\mathrm{PTQ}_{\mathcal B}$ from scratch on the identical
calibration tensor and finally optimize a fresh $\Phi_{\mathrm{post}}$ while
freezing the second quantized graph. Thus the complete method is
$\Phi_{\mathrm{pre}}\!\rightarrow\!\mathrm{PTQ}_{\mathcal B}
\!\rightarrow\!\Phi_{\mathrm{post}}$. The one-sided appendix results instead
apply only $\mathrm{PTQ}_{\mathcal B}\!\rightarrow\!\Phi_{\mathrm{post}}$.
Budgets are matched \emph{per affine stage}; the complete method contains two
200-step affine stages and a second PTQ pass.

\paragraph{Affine optimization.}
Unless stated otherwise, each stage uses 200 AdamW steps, batch size 1, sequence
length 256, learning rate $5\times10^{-4}$, zero weight decay, no learning-rate
schedule, and gradient-norm clipping at 1.0. Training minimizes equal-weight
cross entropy and full-vocabulary teacher distillation at temperature 1. The
teacher is the original dense model, evaluated with \texttt{use\_cache=False};
only normalization-affine tensors receive gradients. Calibration subblocks are
sampled deterministically (seed 100003), and checkpoints are written every ten
steps. For AWQ, relative log-scales use learning rate $10^{-4}$ and are clamped
to $[-0.1,0.1]$. No channel-scale regularizer is used in the LLM experiments.

\paragraph{Evaluation.}
WikiText-2 perplexity uses the complete test split in non-overlapping 2,048-token
blocks. C4 perplexity uses the first 1,100 validation documents and the first 256
full 2,048-token blocks. Zero-shot evaluation uses lm-eval-harness 0.4.9.1,
maximum length 2,048, and batch size 4. We report normalized accuracy for PIQA,
ARC-Easy, ARC-Challenge, and HellaSwag, and accuracy for WinoGrande and BoolQ;
Avg. is their unweighted mean. The software stack uses Transformers 4.54.1,
Datasets 3.6.0, and Accelerate 1.9.0. Quantization and evaluation were executed
on MetaX C500 and C600-A accelerators. Run manifests record model, calibration,
rotation-artifact, software, and hardware identifiers.

\paragraph{Baseline provenance.}
Unless explicitly attributed to prior work, reported baselines and paired
affine variants are rerun in our evaluation stack from the same dense
checkpoint and tokenizer. Within a model--regime pair, they share the dataset
snapshots, calibration indices, preprocessing, and metric implementation.
Tables that reproduce values from a source paper identify that provenance in
their captions; run manifests record the code revision and generated artifact
used for every unmarked result.

\paragraph{Construction cost.}
Table~\ref{tab:sandwich_cost} separates the two PTQ executions from the two
short affine stages.  Evaluation is excluded because it is identical across
the compared construction variants and often dominates end-to-end job time.

\begin{table*}[t]
\centering
\caption{\textbf{Representative offline construction cost of \sandwichquant.}
Wall-clock time is decomposed into the disposable PTQ probe, pre-PTQ affine
adaptation, the deployable PTQ rebuild, and post-PTQ affine adaptation.
Times are reconstructed from representative single-accelerator C500/C600-A
runs and include graph construction and checkpoint I/O but exclude downstream
evaluation.  The two affine stages update only $|\Phi|$ native normalization
parameters.  The final graph introduces no additional parameters or inference
operators relative to its base PTQ graph.}
\label{tab:sandwich_cost}
{\small
\setlength{\tabcolsep}{3.8pt}
\renewcommand{\arraystretch}{1.08}
\resizebox{\textwidth}{!}{%
\begin{tabular}{@{}lllrrrrrrr@{}}
\toprule
\textbf{Model} & \textbf{Quantization setting} & \textbf{Device}
& $|\Phi|$
& \multicolumn{4}{c}{\textbf{Wall-clock time (min)}}
& \textbf{Total}
& \textbf{Peak mem. (GiB)} \\
\cmidrule(lr){5-8}
& & &
& \textbf{Probe PTQ}
& $\boldsymbol\Phi_{\rm pre}$
& \textbf{Deploy PTQ}
& $\boldsymbol\Phi_{\rm post}$
& \textbf{(min)} & \\
\midrule
Llama3-8B & GPTQ W3A16, g128
& C600-A & 266,240 & 67.2 & 1.3 & 67.2 & 1.3 & 137.0 & 16.1 \\
Llama3-8B & QuaRot+ResComp W2A4KV4, g128
& C600-A & 266,240 & 89.0 & 1.4 & 89.0 & 1.4 & 180.8 & 25.0 \\
Qwen3-8B & GPTQ W3A16, g128
& C500 & 299,008 & 55.0 & 1.6 & 55.0 & 1.6 & 113.2 & 17.0 \\
Qwen3-8B & QuaRot+ResComp W2A4KV4, g128
& C500 & 299,008 & 85.0 & 1.6 & 85.0 & 1.6 & 173.2 & 25.0 \\
\bottomrule
\end{tabular}%
}}
\end{table*}

\paragraph{Order ablation.}
The four rows in Table~\ref{tab:order_ablation} reuse the same dense checkpoint,
backend configuration, calibration tensor, and seed. ``PTQ'' performs no affine
update; $\mathrm{PTQ}\!\rightarrow\!\Phi_{\mathrm{post}}$ performs one
200-step update after freezing PTQ; $\Phi_{\mathrm{pre}}\!\rightarrow\!
\mathrm{PTQ}$ transfers the first affine solution to the dense model and reruns
PTQ without the second affine update; and \sandwichquant{} applies both
independently initialized 200-step stages. Consequently, compute is matched per
stage rather than across rows with different numbers of stages.

\paragraph{Matched-budget parameter subspaces.}
The Qwen3-8B W2A4KV4 control fixes the completed QuaRot--ResComp graph and gives
each trainable branch exactly $|\Phi|=299{,}008$ scalar coordinates. RandomW
uses a fixed uniformly sampled weight mask; TopGradW retains the coordinates
with the largest calibration-gradient magnitude; AttentionW and MLPW restrict
the same budget to attention and feed-forward projections, respectively.
Masks are selected once and held fixed. All non-frozen branches use the same
calibration blocks, CE+KD objective, 200 updates, batch size, sequence length,
optimizer, and per-stage learning-rate selection protocol as the affine branch.
The frozen branch performs no update. This comparison tests coordinate
structure, not unconstrained capacity or separately tuned asymptotic optima.

\paragraph{Mechanism probes.}
All four-state and structured-source measurements use 16 sequences of length
512 with token stride 16. The four-state intervention evaluates the dense and
quantized graphs at $\Phi_0$ and $\Phi_*$ against a shared dense reference.
Task-aware geometry reports common-reference logit NMSE, teacher KL, a
teacher-Fisher quadratic form, true-token margin NMSE, and the full NLL gap.
Structured-source interventions enable weight, activation, key-cache, and
value-cache perturbations separately and jointly while holding inputs,
references, and affine states fixed. Reported recovery is
$1-E(\Phi_*)/E(\Phi_0)$; negative values therefore denote an increase in that
particular local discrepancy, not a failure of end-to-end recovery.

\paragraph{SpinQuant extension.}
SpinQuant experiments use the released model-specific W16A4KV4 rotation
checkpoint as the fixed rotation initialization, followed by the same W2
weight quantization, 4-bit activation/KV graph, and two affine stages described
above. The exact rotation filename and checksum are stored with each run
manifest; no rotation parameter is updated by either affine stage.

\section{Supplementary Mechanism Derivations and Limits}
\label{app:mechanism_derivation}
\deferredmechanism

\section{Vision and QAT Extensions}
\label{app:vision_qat}
\deferredqatmethod

\paragraph{Quantization graph.}
The CNN experiments use MQBench~\cite{li2021mqbench}. Weights are quantized
symmetrically per output channel; activations are quantized asymmetrically per
tensor with MSE observers and \texttt{FixedFakeQuantize}. ImageNet and
Cityscapes quantize the full network. In the CIFAR-100 checkpoint diagnostic,
the first and last quantized layers remain at 8 bits while internal layers use
the W/A precision stated in the table. Every comparison reloads an identical
frozen activation-observer state before optimization.

Some vision controls additionally use the channel-scale regularizer
\begin{equation}
\mathcal L_{\rm CS}=\frac{1}{L}\sum_{l=1}^{L}
\operatorname{Var}_{c}\!\left[
\log\big(|\gamma_{l,c}|+\epsilon\big)\right],
\label{eq:channel_scale_reg}
\end{equation}
which discourages a few channels from dominating the activation range. This
term is used only where stated below; all headline LLM experiments set its
coefficient to zero.

\paragraph{CIFAR-100 post-QAT adaptation.}
For each source checkpoint, \sqpost{} freezes all convolutional, linear, and
quantizer parameters and updates only BatchNorm scales and biases. The
optimization set contains 40K images and a fixed 4K subset is used for model
selection. All trainable variants use the same CE+KD objective, AdamW with
learning rate $10^{-4}$ and zero weight decay, cosine annealing for 20 epochs,
temperature $T=4$, KD weight 1, and gradient-norm clipping at 1.0. The FP
teacher and image order are shared across paired runs.

\paragraph{ImageNet and Cityscapes PTQ controls.}
RTN applies the calibrated fake-quantization graph without reconstruction.
QDrop~\cite{qdrop} reconstructs quantized blocks using its released stochastic
quantization-drop objective. \sqpost{} optimizes only the normalization affine
coordinates after calibration. The reported ``\sqpre{}+QDrop'' rows transfer
only the learned affine state to the dense checkpoint and then rerun QDrop on
the same calibration samples; no convolutional or linear weight is updated by
the affine stage. ImageNet uses top-1 evaluation at the standard validation
resolution. Cityscapes follows the U-Net crop, normalization, and validation
protocol of the source checkpoint. All paired rows share calibration images,
preprocessing, fake-quantization modules, and evaluation code.

\paragraph{CNN optimization.}
For ImageNet, standalone \sqpost{} is tuned for 20 epochs on the full training
set, and backend-aligned \sqpre{} before QDrop is tuned for 10 epochs. Both use
AdamW with learning rate $5\times10^{-4}$, zero weight decay, batch size 128,
distillation temperature $T=4$, $\lambda_{\rm KD}=1$, and gradient clipping at
1.0; Eq.~\eqref{eq:channel_scale_reg} uses coefficient $10^{-4}$. BN running
statistics are frozen. Quantizer calibration uses 20 mini-batches and at most
5K images; observers are recalibrated after each tuning epoch. QDrop uses 32
calibration batches, 20K reconstruction iterations, stochastic drop
probability 0.5, scale learning rate $4\times10^{-5}$, warm-up ratio 0.2,
rounding weight 0.01, temperature range $[20,2]$, and learned hard-sigmoid
rounding. The CIFAR-100 source QAT stage uses 100 epochs of SGD (momentum 0.9,
learning rate $10^{-3}$, weight decay $5\times10^{-4}$, cosine decay); its
\sqpost{} stage uses 20 epochs at learning rate $10^{-4}$. Cityscapes uses
$512\times512$ crops, batch size 4, and 128 fixed calibration images.

\paragraph{Scope.}
These vision results are controls for cross-architecture generality, not the
headline formulation of \sandwichquant{}. The main LLM tables use the complete
$\Phi_{\mathrm{pre}}\!\rightarrow\!\mathrm{PTQ}\!\rightarrow\!
\Phi_{\mathrm{post}}$ pipeline, whereas the vision rows explicitly identify
whether the affine intervention is post-PTQ or pre-reconstruction.

\section{Controls, Diagnostics, and Practical Boundaries}
\label{app:controls}
\deferredcontrols

\subsection{Data, Subspace, and Negative Controls}
\label{app:llm_diagnostics}

\begin{table}[H]
\centering
\begin{minipage}[t]{0.37\textwidth}
\vspace{0pt}
\centering
\resizebox{\linewidth}{!}{%
\setlength{\tabcolsep}{1.7pt}
\renewcommand{\arraystretch}{1.02}
\begin{tabular}{@{}llcccc@{}}
\toprule
\multirow{2}{*}{\textbf{Model}}
& \multirow{2}{*}{\textbf{Bits}}
& \multirow{2}{*}{\textbf{FP32}}
& \multicolumn{3}{c}{\textbf{Tuning Samples}} \\
\cmidrule(lr){4-6}
& & & \textbf{5K} & \textbf{10K} & \textbf{Full} \\
\midrule
\multirow{2}{*}{MobileNetV2}
& W4A4 & \multirow{2}{*}{71.87}
& 2.98 & 3.88 & \textbf{66.11} \\
& W3A3 &
& 0.09 & 0.11 & \textbf{40.21} \\
\bottomrule
\end{tabular}
}
\captionof{table}{Effect of affine-stage tuning-set size on ImageNet MobileNetV2.
Results are top-1 accuracy (\%); FP32 denotes the full-precision baseline.}
\label{tab:data_dependence}
\end{minipage}
\hfill
\begin{minipage}[t]{0.60\textwidth}
\vspace{0pt}
\centering
\resizebox{\linewidth}{!}{%
\setlength{\tabcolsep}{2.1pt}
\renewcommand{\arraystretch}{1.02}
\begin{tabular}{@{}lcccc@{}}
\toprule
\textbf{Subset}
& \textbf{Params.}
& \textbf{W4A4}
& \textbf{W3A3}
& \textbf{Time} \\
\midrule
RTN & 0 & 0.33 & 0.09 & -- \\
Random equal-size & 34.2K & 2.56 & 0.21 & 0.98$\times$ \\
Top-gradient equal-size & 34.2K & 12.48 & 2.67 & 1.07$\times$ \\
Top-Fisher equal-size & 34.2K & 18.35 & 4.92 & 1.22$\times$ \\
Classifier subset & 34.2K & 0.81 & 0.12 & 0.94$\times$ \\
NoNorm & 3.47M & 57.42 & 31.68 & 1.71$\times$ \\
\rowcolor{oursrow}
\sandwichquant{} & 34.2K & \textbf{66.11} & \textbf{40.21} & 1.00$\times$ \\
Full tuning & 3.52M & 64.28 & 38.74 & 1.83$\times$ \\
\bottomrule
\end{tabular}
}
\captionof{table}{Matched parameter-subspace comparison on ImageNet
MobileNetV2. Time is normalized to the affine branch and includes parameter-selection
overhead.}
\label{tab:parameter_subspace}
\end{minipage}
\end{table}

\section{Rotation-Backend Extension}
\label{app:spinquant}
To test whether the two-stage affine intervention depends on a fixed rotation,
we replace QuaRot by model-specific learned SpinQuant rotations while retaining
the same SandwichQuant state transitions.  Table~\ref{tab:spinquant_sandwich}
reports this backend extension; the rotation artifact identifiers and hashes
are recorded in the corresponding run manifests.

\begin{table}[H]
\centering
\caption{\textbf{SpinQuant-based SandwichQuant results under W2A4KV4.}
Highlighted rows apply the complete two-stage pipeline with the model-specific
learned rotation used by the corresponding SpinQuant baseline. Artifact and
rotation identifiers are stored in the run manifests.}
\label{tab:spinquant_sandwich}

\resizebox{\textwidth}{!}{%
\begin{tabular}{c|l|rr|rrrrrr|c}
\hline\hline
\multicolumn{1}{c|}{\multirow{2}{*}{\textbf{Model}}}
& \multicolumn{1}{c|}{\multirow{2}{*}{\textbf{Method}}}
& \multicolumn{2}{c|}{\textbf{Perplexity $\downarrow$}}
& \multicolumn{6}{c|}{\textbf{Zero-shot Accuracy (\%) $\uparrow$}}
& \multirow{2}{*}{\textbf{Avg. $\uparrow$}} \\
\cline{3-10}
& & \textbf{Wiki2} & \textbf{C4}
& \textbf{PIQA} & \textbf{ARC-E} & \textbf{ARC-C}
& \textbf{HS} & \textbf{WG} & \textbf{BoolQ} & \\
\hline

& FP16 & 5.47 & 7.27 & 78.9 & 74.6 & 46.1 & 75.9 & 69.2 & 77.7 & 70.4 \\
\cline{2-11}
& SpinQuant + GPTAQ & 11.6 & 26.5  & 62.2  & 42.3  & 25.5 & 40.9 & 54.7 & \textbf{63.4} & 48.1 \\
\rowcolor{oursrow}
\cellcolor{white} & \sandwichquant{} (SpinQuant+GPTAQ) & 8.6 & \textbf{15.2}  & 65.6  & \textbf{52.2} & \textbf{30.1} & 53.5 & \textbf{57.6} & 63.1 & \textbf{53.7} \\
& SpinQuant + $\mathrm{ResComp}^{\ast}$ & 11.1  & 24.7  & 62.7  & 45.4  & 27.6 & 42.3 & 55.0 & 62.3 & 49.2 \\
\rowcolor{oursrow}
\cellcolor{white} \multirow{-5}{*}{Llama2-7B} & \sandwichquant{} (SpinQuant+$\mathrm{ResComp}^{\ast}$) & \textbf{8.5}  & 15.3  & \textbf{67.1}  & 50.9 & 27.9 & \textbf{54.0} & 54.7 & 63.2 & 53.0 \\
\hline

& FP16 & 6.14 & 9.45 & 80.9 & 77.7 & 53.2 & 79.2 & 72.9 & 81.2 & 74.2 \\
\cline{2-11}
& SpinQuant + GPTAQ & 18.3 & 55.6 & 61.4 & 42.3 & 26.6 & 40.1 & 54.7 & 62.7 & 47.9  \\
\rowcolor{oursrow}
\cellcolor{white} & \sandwichquant{} (SpinQuant+GPTAQ) & 14.0 & \textbf{32.6}  & \textbf{63.6} & 47.1 & 27.8 & \textbf{50.4} & \textbf{56.8} & \textbf{65.9} & \textbf{51.9} \\
& SpinQuant + $\mathrm{ResComp}^{\ast}$ & 18.1 & 53.7 & 60.6 & 41.4 & 28.1 & 40.8 & 54.1 & 62.1 & 47.9  \\
\rowcolor{oursrow}
\cellcolor{white} \multirow{-5}{*}{Llama3-8B} & \sandwichquant{} (SpinQuant+$\mathrm{ResComp}^{\ast}$)  & \textbf{13.1} & 34.3 & 61.3 & \textbf{47.2} & \textbf{28.9} & 49.7 & 55.7 & 64.3 & 51.2  \\
\hline
\end{tabular}}
\end{table}

\section{One-Sided Post-Quantization Correction}
\label{app:llm_details}
The headline results use the complete \sandwichquant{} pipeline. To isolate
its second stage, Tables~\ref{tab:llm_w3_post_only} and
\ref{tab:llm_w2a4kv4_post_only} instead freeze an already completed PTQ graph
and optimize only $\Phi_{\mathrm{post}}$. These are component results, not an
alternative definition of our method.
\begin{table}[H]
\centering
\caption{\textbf{One-sided \sqpost{} results under W3A16 (group size 128).}
Each highlighted row freezes the completed backend in the preceding row and
optimizes only $\Phi_{\mathrm{post}}$. These component results are intentionally
separated from the full \sandwichquant{} rows in Table~\ref{tab:llm_w3_main}.}
\label{tab:llm_w3_post_only}

\resizebox{\textwidth}{!}{%
\begin{tabular}{c|l|rr|rrrrrr|c}
\hline\hline
\multicolumn{1}{c|}{\multirow{2}{*}{\textbf{Model}}}
& \multicolumn{1}{c|}{\multirow{2}{*}{\textbf{Method}}}
& \multicolumn{2}{c|}{\textbf{Perplexity $\downarrow$}}
& \multicolumn{6}{c|}{\textbf{Zero-shot Accuracy (\%) $\uparrow$}}
& \multirow{2}{*}{\textbf{Avg. $\uparrow$}} \\
\cline{3-10}
& & \textbf{Wiki2} & \textbf{C4}
& \textbf{PIQA} & \textbf{ARC-E} & \textbf{ARC-C}
& \textbf{HS} & \textbf{WG} & \textbf{BoolQ} & \\
\hline

& FP16 & 5.47 & 7.27 & 78.9 & 74.6 & 46.1 & 75.9 & 69.2 & 77.7 & 70.4 \\
\cline{2-11}
& RTN & 8.22 & 11.53  & 76.3  & 66.5  & 40.6 & 68.7 & 65.8 & 67.5 & 64.2 \\
\rowcolor{oursrow}
\cellcolor{white} & RTN + \sqpost{} & 7.47  &  10.42 & 77.3  & 68.9  & 41.7 & 70.9 & 66.0 & 70.8 & 65.9  \\
& AWQ & 6.79  & 8.93  & 77.1  & 69.9  & 41.8 & 71.2 & \textbf{67.7} & 71.5 & 66.3 \\
\rowcolor{oursrow}
\cellcolor{white} & AWQ + \sqpost{} &  6.53 & 8.71  &  77.2 & \textbf{70.7}  & \textbf{43.0} & 72.4 & 67.2 & 72.8 & \textbf{67.2} \\
& GPTQ & 6.75 & 13.72 & 76.8 & 66.5 & 39.7 & 68.3 & 67.5 & 68.9 & 64.6 \\
\rowcolor{oursrow}
\cellcolor{white} & GPTQ + \sqpost{} & 6.31 & 8.33 & 77.2 & 66.3 & 41.6 & 72.2 & 67.6 & 69.3 & 65.7 \\
& GPTAQ &  6.81 & 8.40  & 77.8  & 69.2  & 40.7 & 71.9 & 67.6 & 71.5 & 66.5 \\
\rowcolor{oursrow}
\cellcolor{white} & GPTAQ + \sqpost{} & 6.31  & 8.18  & \textbf{78.2}  & 70.5  & 40.9 & 72.0 & 66.6 & 71.2 & 66.6 \\
& $\mathrm{ResComp}^{\ast}$ & 6.25  & 8.19  & 77.4  & 69.5  & 41.5 & 72.3 & 67.4 & 71.5 & 66.6 \\
\rowcolor{oursrow}
\cellcolor{white} \multirow{-11}{*}{Llama2-7B} & $\mathrm{ResComp}^{\ast}$ + \sqpost{} & \textbf{6.22}  &  \textbf{8.16} & 77.5  &  69.4 & 42.3 & \textbf{72.7} & 67.0 & \textbf{73.2} & 67.0 \\
\hline

& FP16 & 6.14 & 9.45 & 80.9 & 77.7 & 53.2 & 79.2 & 72.9 & 81.2 & 74.2 \\
\cline{2-11}
& RTN & 29.21 & 43.1 & 68.5 & 50.1 & 35.4 & 53.5 & 58.8 & 63.7 & 55.0 \\
\rowcolor{oursrow}
\cellcolor{white} & RTN + \sqpost{} & 15.2 & 22.3 & 73.1 & 57.1 & 35.5 & 63.2 & 63.2 & 66.9 & 59.8 \\
& AWQ & 9.53 & 14.74 & 76.1 & 69.2 & 42.2 & 71.4 & 69.0 & 78.2 & 67.7 \\
\rowcolor{oursrow}
\cellcolor{white} & AWQ + \sqpost{} & 9.09 & 14.10 & 77.6 & 71.1 & 44.2 & 71.4 & 70.1 & 76.5 & 68.5 \\
& GPTQ & 8.33 & 13.11 & 77.5 & 70.6 & 43.7 & 71.9 & 71.4 & 77.9 & 68.8 \\
\rowcolor{oursrow}
\cellcolor{white} & GPTQ + \sqpost{} & 7.89 & 12.21 & 78.2 & 72.1 & 44.5 & 74.4 & 72.3 & \textbf{79.1} & 70.1 \\
& GPTAQ &  8.13 & 12.77  & 77.3  & 69.3  & 43.6 & 61.9 & 72.0 & 77.7 & 67.0 \\
\rowcolor{oursrow}
\cellcolor{white} & GPTAQ + \sqpost{} & 7.97  & 12.36  & 77.2  &  69.8 & 44.3 & 67.6 & 72.4 & 78.1 & 68.2 \\
& $\mathrm{ResComp}^{\ast}$ & 7.77 & 12.25 & 77.7 & 73.8 & 45.7 & 74.6 & 72.3 & \textbf{79.1} & 70.5 \\
\rowcolor{oursrow}
\cellcolor{white} \multirow{-11}{*}{Llama3-8B} & $\mathrm{ResComp}^{\ast}$ + \sqpost{} & \textbf{7.65} & \textbf{12.17} & \textbf{79.2} & \textbf{75.1} & \textbf{47.9} & \textbf{75.1} & \textbf{72.5} & 77.1 & \textbf{71.2} \\
\hline

& FP16 & 9.72 & 15.40 & 77.8 & 80.9 & 56.7 & 74.9 & 67.8 & 86.6 & 74.1 \\
\cline{2-11}
& RTN & 23.54 & 33.86 & 68.2 & 58.7 & 35.3 & 54.6 & 54.9 & 63.9 & 55.9 \\
\rowcolor{oursrow}
\cellcolor{white} & RTN + \sqpost{} & 17.06 & 24.86 & 73.0 & 67.3 & 43.2 & 63.7 &62.1 & 76.4 & 64.3 \\
& AWQ & 12.11 & 18.49 & 74.3 & 71.3 & 45.8 & 67.8 & 63.3 & 82.5 & 67.5 \\
\rowcolor{oursrow}
\cellcolor{white} & AWQ + \sqpost{} & 11.53 & 17.64 & 74.8 & 71.8 & 46.9 & 68.4 & 63.9 & 82.8 & 68.1 \\
& GPTQ & 11.29 & 17.09 & 76.9 & 76.8 & 50.9 & 71.6 & \textbf{69.2} & \textbf{85.8} & 71.9 \\
\rowcolor{oursrow}
\cellcolor{white} & GPTQ + \sqpost{} & \textbf{10.59} & \textbf{16.38} & 76.8 & 76.9 & 51.9 & \textbf{72.4} & \textbf{69.2} & \textbf{85.8} & 72.2 \\
& GPTAQ & 11.37  & 17.13  & 75.7  & 72.2  & 48.7 & 70.9 & 67.9 & 85.7 & 70.2 \\
\rowcolor{oursrow}
\cellcolor{white} & GPTAQ + \sqpost{} & 10.88  & 16.61  & 75.9  & 71.3  & 48.9 & 71.3 & 68.7 & 85.2 & 70.2 \\
& $\mathrm{ResComp}^{\ast}$ & 11.66  & 17.45  & 76.6  & 77.6  & 53.2 & 70.7 & 67.3 & 84.7 &  71.7 \\
\rowcolor{oursrow}
\cellcolor{white} \multirow{-11}{*}{Qwen3-8B} & $\mathrm{ResComp}^{\ast}$ + \sqpost{} &  11.02 &  16.78 & \textbf{77.3}  &  \textbf{78.2} & \textbf{54.1} & 71.4 & 67.8 & 84.8 & \textbf{72.3} \\
\hline\hline
\end{tabular}}
\end{table}

\begin{table}[H]
\centering
\caption{\textbf{One-sided \sqpost{} results under joint W2A4KV4 (group size 128).}
Each highlighted row freezes the completed QuaRot backend and optimizes only
$\Phi_{\mathrm{post}}$. Calibration uses 128 WikiText-2 sequences of length
2,048 (seed 0).}
\label{tab:llm_w2a4kv4_post_only}

\resizebox{\textwidth}{!}{%
\begin{tabular}{c|l|rr|rrrrrr|c}
\hline\hline
\multicolumn{1}{c|}{\multirow{2}{*}{\textbf{Model}}}
& \multicolumn{1}{c|}{\multirow{2}{*}{\textbf{Method}}}
& \multicolumn{2}{c|}{\textbf{Perplexity $\downarrow$}}
& \multicolumn{6}{c|}{\textbf{Zero-shot Accuracy (\%) $\uparrow$}}
& \multirow{2}{*}{\textbf{Avg. $\uparrow$}} \\
\cline{3-10}
& & \textbf{Wiki2} & \textbf{C4}
& \textbf{PIQA} & \textbf{ARC-E} & \textbf{ARC-C}
& \textbf{HS} & \textbf{WG} & \textbf{BoolQ} & \\
\hline

& FP16 & 5.47 & 7.27 & 78.9 & 74.6 & 46.1 & 75.9 & 69.2 & 77.7 & 70.4 \\
\cline{2-11}
& QuaRot + GPTAQ & 11.7 & 24.8 & 62.3 & 45.6 & 25.6 & 41.2 & 54.0 & 62.4 & 48.5 \\
\rowcolor{oursrow}
\cellcolor{white} & QuaRot + GPTAQ + \sqpost{} & 8.4 & \textbf{14.6} & 66.2 & 49.5 & \textbf{30.5} & \textbf{54.7} & 57.5 & 63.3 & 53.6 \\
& QuaRot + $\mathrm{ResComp}^{\ast}$ & 11.5 & 23.6 & 63.6 & 46.4 & 24.9 & 40.8 & 56.0 & 61.9 & 48.9 \\
\rowcolor{oursrow}
\cellcolor{white}\multirow{-5}{*}{Llama2-7B} & QuaRot + $\mathrm{ResComp}^{\ast}$ + \sqpost{} & \textbf{8.3} & 14.7 & \textbf{66.6} & \textbf{50.2} & 29.1 & 54.3 & \textbf{58.2} & \textbf{64.9} & \textbf{53.9}  \\
\hline

& FP16 & 6.14 & 9.45 & 80.9 & 77.7 & 53.2 & 79.2 & 72.9 & 81.2 & 74.2 \\
\cline{2-11}
& QuaRot + GPTAQ & 23.0 & 62.8 & 55.8 & 37.2 & 22.3 & 36.5 & 50.4 & 59.3 & 43.6 \\
\rowcolor{oursrow}
\cellcolor{white} & QuaRot + GPTAQ + \sqpost{} & 12.9 & \textbf{29.9} & 63.2 & \textbf{49.2} & \textbf{28.7} & \textbf{50.8} & 56.9 & 64.7 & 52.3 \\
& QuaRot + $\mathrm{ResComp}^{\ast}$ & 22.1 & 61.5 & 56.0 & 38.5 & 24.6 & 35.8 & 53.5 & 60.9 & 44.9 \\
\rowcolor{oursrow}
\cellcolor{white} \multirow{-5}{*}{Llama3-8B} & QuaRot + $\mathrm{ResComp}^{\ast}$ + \sqpost{} & \textbf{12.8} & 30.3 & \textbf{64.7} & 48.4 & 28.3 & 50.2 & \textbf{57.7} & \textbf{65.6} & \textbf{52.5} \\
\hline

& FP16 & 9.72 & 15.40 & 77.8 & 80.9 & 56.7 & 74.9 & 67.8 & 86.6 &  74.1\\
\cline{2-11}
& QuaRot + GPTAQ & 28.2 & 64.1 & 57.9 & 39.5 & 23.4 & 34.7 & 51.7 & 61.6 & 44.8 \\
\rowcolor{oursrow}
\cellcolor{white} & QuaRot + GPTAQ + \sqpost{} & 15.0 & 33.5 & 65.5 & 51.4 & 29.5 & 46.3 & 57.1 & 66.6 & 52.7 \\
& QuaRot + $\mathrm{ResComp}^{\ast}$ & 18.7 & 38.3 & 61.6 & 44.4 & 28.6 & 43.7 & 54.8 & 64.4 & 49.6 \\
\rowcolor{oursrow}
\cellcolor{white} \multirow{-5}{*}{Qwen3-8B} & QuaRot + $\mathrm{ResComp}^{\ast}$ + \sqpost{} & \textbf{12.7} & \textbf{26.8} & \textbf{66.9} & \textbf{56.1} & \textbf{34.5} & \textbf{52.5} & \textbf{59.2} & \textbf{69.7} & \textbf{56.5} \\

\hline\hline
\end{tabular}}
\end{table}

\end{document}